\documentclass[3p]{elsarticle}

\usepackage{siunitx}
\usepackage{stackengine}
\stackMath
\newcommand\tenbar[2][1]{%
  \def\useanchorwidth{T}%
  \ifnum#1>1
    \stackunder[0.5pt]{\tenbar[\numexpr#1-1\relax]{#2}}{\scriptscriptstyle\rule{1.4ex}{0.6pt}}%
  \else
    \stackunder[2pt]{#2}{\scriptscriptstyle\rule{1.4ex}{0.6pt}}%
  \fi
}

\usepackage{hyperref}
\hypersetup{
    colorlinks,
    linkcolor={magenta!80!black},
    citecolor={blue!80!black},
    urlcolor={blue!80!black}
}

\usepackage{algorithm}
\usepackage{algorithmic}

\usepackage{enumitem}
\usepackage{physics}
\usepackage{float}
\usepackage{subcaption}

\usepackage[capitalise, noabbrev]{cleveref}

\usepackage{titlesec}
\titleformat{\paragraph}[runin]
  {\normalfont\normalsize\itshape}
  {\theparagraph}{1em}{}

\usepackage{amssymb}
\usepackage{amsmath}
\usepackage{wrapfig}
\usepackage{booktabs}
\usepackage{multirow}

\begin{document}

\begin{frontmatter}



\title{Toward AI-Driven Digital Twins for Metropolitan Floods: A Conditional Latent Dynamics Network Surrogate of the Shallow Water Equations}


\author[label1]{Phillip Si}
\author[label1]{Yuan Qiu}
\author[label2]{Omar Sallam}
\author[label2]{Jeremy Feinstein}
\author[label2]{Ziang He}
\author[label2]{Eugene Yan}
\author[label1]{Peng Chen \corref{cor1}}
\ead{pchen402@gatech.edu}

\affiliation[label1]{organization={School of Computational Science and Engineering, Georgia Institute of Technology},
            addressline={756 West Peachtree Street Northwest}, 
            city={Atlanta},
            postcode={30308}, 
            state={GA},
            country={USA}}

\cortext[cor1]{Corresponding author} 

\affiliation[label2]{organization={Environmental Science Division, Argonne National Laboratory},
            addressline={9700 S Cass Ave}, 
            city={Lemont},
            postcode={60439}, 
            state={IL},
            country={USA}}
            
\begin{abstract}
AI-driven flood digital twins demand fast hydrodynamic surrogates for ensemble forecasting and observation assimilation. Yet even GPU-accelerated two-dimensional shallow water equation (SWE) solvers still require $\sim 55$ minutes per $96$-hour run on a $\sim 4.2$-million-active-cell metropolitan basin (the Des~Plaines River basin at $30\,\mathrm{m}$ resolution), making such workloads prohibitive at native resolution. We present the Conditional Latent Dynamics Network (CLDNet): a low-dimensional latent neural ODE driven by rainfall, paired with a coordinate-based decoder conditioned on static terrain (elevation, slope, Manning roughness) that reconstructs depth and discharge at arbitrary query points. Pointwise decoding decouples memory from grid size and handles irregular watersheds natively, enabling metropolitan-scale training on a single compute node and direct queries at exact gauge coordinates without raster snapping. We evaluate CLDNet on a synthetic $250{,}000$-cell Texas benchmark and on a new Des~Plaines case study of $114$ real-rainfall Stage~IV storms whose reference simulator we validate against United States Geological Survey (USGS) gauges at the April~2013 flood-of-record (Nash--Sutcliffe efficiency $0.57$--$0.94$ on mean-recentered water-surface elevation). CLDNet roughly halves the relative root-mean-squared error of an unconditional baseline, outperforms regular-grid VAE--ConvLSTM and FNO baselines on the Texas benchmark (both presuppose a Cartesian grid and do not apply to the irregular Des~Plaines watershed), reaches a critical success index of $\approx 86\%$ at the $0.5\,\mathrm{m}$ inundation threshold, and produces a full $96$-hour basin-wide forecast in $\sim 29$ seconds---a $\sim 115\times$ speedup.
\end{abstract}



\begin{keyword}
AI-driven digital twins \sep neural surrogate modeling \sep shallow water equations \sep conditional latent dynamics networks \sep terrain conditioning \sep real-time flood forecasting
\end{keyword}

\end{frontmatter}




\section{Introduction}
\label{sec:introduction}

\subsection{Motivation}

Urban and regional flooding is among the most costly and disruptive natural hazards, with $1.81$ billion people, roughly $23\%$ of the world's population, directly exposed to $1$-in-$100$-year flood events~\cite{rentschler2022flood} greater than $0.15\,\mathrm{m}$, and satellite observations document that the proportion of the global population exposed to floods grew by roughly $20$--$24\%$ between 2000 and 2015, an order of magnitude faster than population growth alone~\cite{tellman2021satellite}. Continued urbanization of floodplains and climate-driven intensification of extreme precipitation are projected to amplify these trends further~\cite{tellman2021satellite}. Physics-based hydrodynamic solvers based on the two-dimensional shallow water equations (SWEs), from the LISFLOOD-FP inertial formulation~\cite{bates2010simple} to modern GPU-accelerated Godunov schemes, have become the workhorse for producing high-fidelity flood forecasts at city and basin scale~\cite{xia2017efficient, xia2018new, xia2019full}. Modern GPU implementations such as SynxFlow achieve significant computational efficiency on metropolitan domains, integrating the SWE system at speeds orders of magnitude faster than real time: a single $96$-hour forecast for the $4{,}188{,}840$-active-cell Des Plaines River basin studied in this work completes in $\sim 55$ minutes of GPU wall time at $30\,\mathrm{m}$ resolution (Section~\ref{sec:dataset}). This is already fast enough for a single deterministic forecast, but the cost scales linearly with the number of runs, so workflows that require many forecasts per basin---rainfall-forecast ensembles, what-if and scenario-screening sweeps for flood-risk management, and climate-projection downscaling---remain slow and expensive at native resolution. Closing this gap calls for \emph{fast surrogate models} that (i) reproduce simulator-level accuracy at orders-of-magnitude lower per-trajectory cost, (ii) retain a memory footprint small enough to admit many concurrent inferences on a single GPU, and (iii) scale to metropolitan domains without coarsening or cropping. Delivering all three is the central objective of this work.

Operational flood prediction at metropolitan scale is moving toward \emph{AI-driven digital twins}: living, observation-coupled models of a basin that ingest precipitation, stream-stage, and inundation observations while maintaining a fast representation of the underlying hydrodynamics that can be queried for ensemble forecasts, scenario screens, and water-management decisions. Such a twin requires at least three components: (i) a high-fidelity physical simulator anchored to observations; (ii) a fast surrogate that reproduces the simulator's dynamics at much lower per-run cost so that many forecasts and assimilation cycles can run on routine hardware; and (iii) a compact state representation of the basin that can be updated from sparse observations. This work delivers the second component and part of the third for the Des~Plaines basin: CLDNet provides a fast SWE surrogate with a low-dimensional latent state and a meshless decoder that can be queried at exact gauge coordinates without raster snapping. We position CLDNet as a hydrodynamic surrogate layer \emph{toward} AI-driven flood digital twins, rather than as a complete operational twin; the corresponding online observation-assimilation step is articulated as a contribution-level extension (C5 below) rather than demonstrated here.

\subsection{Data-driven flood surrogates: state of the art and open gaps}

Recent data-driven flood surrogates include five broad families; while each has significantly advanced the state of the art, scaling these approaches to metropolitan areas remains an active area of development.

\emph{Graph neural networks} (GNN) exploit the unstructured nature of hydrodynamic meshes and the local character of SWE flux computations. SWE-GNN~\cite{bentivoglio2023rapid} uses an encoder-processor-decoder GNN with message-passing layers designed to mimic wet-to-dry flux propagation; it was demonstrated on synthetic dike-breach events over Perlin-noise topographies with grids of up to $128\times 128=16{,}384$ cells at $100\,\mathrm{m}$ resolution and generalized to unseen topographies and breach locations. A multi-scale SWE-GNN~\cite{bentivoglio2025multi} (mSWE-GNN) variant extends this formulation with a multi-resolution processor that demonstrates applicability on real topographies. ComGNN~\cite{kazadi2024pluvial} converts the digital elevation model (DEM) into a directed flow-direction graph and runs a two-stage message-passing scheme that respects mass and momentum conservation for pluvial flood emulation. Both methods scale with the number of mesh edges and are challenged by flat urban terrain, backwater effects, and dense metropolitan meshes with millions of nodes. We do not include a GNN baseline in our experiments because (i) ComGNN's two-stage formulation predicts only water depth in a short $200$-minute forecast horizon using $5$-minute time steps, and (ii) message-passing at $4.2$ million active nodes exceeds the GPU memory in current implementations of SWE-GNN and ComGNN.

\emph{Convolutional / U-Net architectures} treat the flood field as an image and benefit from large-receptive-field encoder-decoder backbones. U-FLOOD~\cite{lowe2021uflood} predicts maximum pluvial flood depth maps over the city of Odense using a $\sim 28$M-parameter U-Net, achieving second-scale inference at unseen locations and rain events. However, it predicts only the maximum depth (no temporal dynamics), was designed for a single pluvial catchment, and the memory of its dense decoder scales with the output resolution, limiting deployment on metropolitan meshes.

\emph{Recurrent / hybrid deep learning} stacks a Convolutional Long Short-Term Memory network (ConvLSTM) or similar sequence model on top of a convolutional encoder of the flood state. The Hydro-Y-Net of Yang et al.~\cite{yang2024rapid} couples a physics-informed ConvLSTM to a hydrodynamic model, issuing $12$-step forecasts in $\sim 12$ seconds on a $\sim 250{,}000$-cell Nan'gang River Basin in Guangzhou city. The approach generalizes well within-district, but the ConvLSTM recurrence and dense grid representation have memory footprints that scale with the full spatial grid at every time step, making them impractical to push to the $4.2$ million active-cell scale and beyond.

\emph{Physics-informed neural networks (PINNs)} enforce the SWE residual as a soft constraint on a coordinate Multilayer Perceptron (MLP). The physics-information-fused model of Li et al.~\cite{li2023physical} combines a PINN with a subregion-specific collocation sampling scheme that concentrates residual points near high-gradient regions, achieving orders-of-magnitude speedups over 2D SWE solvers on dam-break benchmarks. A broader body of work on PINN-based SWE solvers has emerged in recent years~\cite{bihlo2022physics, brecht2025physics}. However, standard coordinate-input PINNs must be retrained per geometry and per event, do not accept rainfall or DEM as distributed input fields, and have not been demonstrated on realistic rainfall-driven urban domains.

\emph{Neural operators} learn the solution map of a parametric partial differential equation (PDE) family directly. The Fourier Neural Operator (FNO)~\cite{li2020fourier} parameterizes an integral kernel in Fourier space; DeepONet~\cite{lu2021learning} uses a branch--trunk decomposition and is a universal approximator of operators between Banach spaces; physics-informed variants further constrain the operator with PDE residuals~\cite{li2024physics}. In the data-limited regime, sample efficiency on high-dimensional parameter inputs can be improved by adding a parameter-Jacobian loss, as in derivative-informed neural operators~\cite{o2022derivative,olearyroseberry2024dino}, the derivative-enhanced DeepONet variant~\cite{qiu2024dedeeponet}, and the latent attention neural operator that fuses derivative-informed dimension reduction with attention in latent space~\cite{go2025lano}; reduced-basis neural operator training with a first-order system least-squares objective yields a loss provably equivalent to the PDE-induced solution error norm, thus serving as a rigorous a posteriori error estimator~\cite{varopl2025}. In this work, we select FNO as a representative baseline for the Texas dataset (Section~\ref{sec:baselines}), as in~\cite{sun2023rapid}, owing to the flexibility of its conditioning mechanism. The global spectral parameterization that gives FNO its sample efficiency on smooth problems is, however, challenged by the sharp wet/dry interfaces of metropolitan flood fields (Section~\ref{sec:baselines_discussion}).

Against this digital-twin target, existing flood surrogates still leave several requirements underserved. Fast SWE surrogates such as the one developed here can provide the hydrodynamic-replication layer of a flood digital twin, but the surrogate must also scale to operational domains, preserve terrain-controlled temporal dynamics, and expose a state/query interface that can ultimately be connected to sparse observations. This view is consistent with recent hydrological digital-twin studies, which emphasize the same three-pillar structure---physics-anchored simulation, hybrid physics + AI dynamics replication, and real-time data assimilation from gauges and remote sensing---in support of water-system monitoring and management decisions~\cite{brocca2024dthydrology, yang2024digitaltwinriverbasins, rapalo2024floodforecastingdigitaltwin, nguyen2025alzette}.

A recent comprehensive review~\cite{liu2025comprehensive} catalogues over $50$ machine-learning flood-depth estimators and confirms the picture above: the largest reported demonstrations for end-to-end neural flood surrogates remain well below metropolitan scale (Table~\ref{tab:flood_problem_sizes}). Three requirements for operational flood digital twins remain underserved:

\begin{enumerate}
    \item \textbf{Metropolitan scalability.} Existing surrogates have been demonstrated on domains mostly below metropolitan/million-cell scale cells, roughly two orders of magnitude smaller than a metropolitan watershed at $30\,\mathrm{m}$ resolution. Memory costs in grid-based architectures (GNN, CNN, U-Net, ConvLSTM, FNO) typically scale with the full grid size, which forces coarsening or cropping of operationally realistic domains.
    \item \textbf{Terrain-aware temporal dynamics validated against real events.} Flood dynamics are jointly governed by time-varying rainfall and spatially heterogeneous static terrain (elevation, slope, roughness). Most published surrogates either predict static inundation maxima (losing dynamics) or handle terrain only implicitly through the convolutional inductive bias, and event-level validation against real gauge observations is rare. All five representative surrogates in Table~\ref{tab:flood_problem_sizes} are in fact \emph{simulator-only}: their ML targets are outputs of a companion hydrodynamic solver: Delft3D-FM for SWE-GNN~\cite{bentivoglio2023rapid}, MIKE~21 for U-FLOOD~\cite{lowe2021uflood}, a 2-D SWE code for Li et al.~\cite{li2023physical}, an internal 2-D hydrodynamic model for Hydro-Y-Net~\cite{yang2024rapid}, and LISFLOOD-FP for ComGNN~\cite{kazadi2024pluvial}. None of them compare predicted depth, stage, or discharge to observations via gauge time series, high-water marks, or remote-sensing extents. Hydro-Y-Net ingests real gauge water level as an \emph{input} but not as a validation target. Consequently, none of these demonstrations provides an end-to-end accuracy bound anchored to real-world observations at gauge cross-sections.
    \item \textbf{Observation-compatible state and query interface.} A flood digital twin must eventually update its basin state from sparse gauge and remote-sensing observations. Most grid-based surrogates expose only raster outputs tied to the simulation mesh, requiring nearest-neighbor or interpolation steps for sensor-coordinate queries and off-grid cross-section evaluation, and providing no compact latent state amenable to data assimilation. A surrogate intended as a digital-twin component should therefore provide both a compact dynamical state and a flexible, mesh-free observation/query operator.
\end{enumerate}

\subsection{Contributions of this work}

We address these gaps by developing and evaluating a \emph{conditional latent dynamics network} (CLDNet) surrogate for SWE-based regional flood forecasting. Our starting point is the Latent Dynamics Network (LDNet) of Regazzoni et al.~\cite{regazzoni2024learning}, a meshless surrogate paradigm that factorizes a spatio-temporal PDE solution into (i) a low-dimensional latent neural dynamical system driven by time-varying inputs and (ii) a coordinate-based reconstruction MLP that decodes the field at arbitrary query points. LDNets were originally demonstrated on synthetic PDE benchmarks (advection-diffusion, Navier-Stokes, cardiac electrophysiology) on non-dimensional unit-square domains with no terrain or land-surface forcing. We show that, with the right conditioning mechanism and a training recipe designed around the coordinate-based decoder, this paradigm scales to real regional flood modeling at metropolitan resolution. Concretely, contributions C1 and C2 address the metropolitan-scalability gap (1) by combining a meshless decoder with terrain conditioning; contributions C2 and C3 address the terrain-dynamics-and-real-event-validation gap (2) through the terrain-aware decoder and the USGS-validated Des~Plaines benchmark; and contribution C5 addresses the observation-interface gap (3) by exposing a compact, gauge-queryable latent state, with C4 providing the multi-metric evaluation that ties them together.

Concretely, our contributions are:

\begin{enumerate}
    \item[\textbf{(C1)}] \emph{Meshless decoder at metropolitan scale.} CLDNet's coordinate-based decoder decouples both training and inference memory from the simulation-grid size. Combined with a training recipe built around this property, including spatial subsampling per step and distributed training over events (Section~\ref{sec:training}), the method enables $4{,}188{,}840$-active-cell training on a single dual-GPU node and a $\sim 29\,\mathrm{s}$ full $96$-hour Illinois forecast at inference, a $\sim 115\times$ end-to-end speedup over SynxFlow (Section~\ref{sec:efficiency}). The same model returns a $48$-hour $250{,}000$-cell Texas forecast in $4.3\,\mathrm{s}$.
    \item[\textbf{(C2)}] \emph{Terrain-conditioned decoder.} We augment the LDNet reconstruction network with a static terrain feature vector consisting of standardized elevation, slope magnitude, and scaled Manning roughness, concatenated with the (Fourier-embedded) spatial coordinate at every query point. The conditioning is parameter-cheap and is the single architectural change responsible for the headline accuracy gains ($31.09\% \rightarrow 14.44\%$ test relative root-mean-squared error (rRMSE) on Texas; $35.04\% \rightarrow 17.79\%$ on held-out Illinois events; CSI $= 86.04\%$ at the $0.5\,\mathrm{m}$ threshold on Illinois; Section~\ref{sec:cldnet}, Section~\ref{sec:rmse_results}).
    \item[\textbf{(C3)}] \emph{Metropolitan-scale Des~Plaines benchmark with shape-level USGS water surface elevation (WSE) validation.} We curate a regional-flood dataset of $114$ real-rainfall storm forcings drawn from the NCEP Stage~IV archive (2002--2024) and applied to the Des~Plaines River basin (HUC8~07120004, greater Chicago), with $4{,}188{,}840$ active watershed cells embedded in a $5{,}075\times 1{,}661 = 8{,}429{,}575$ raster at $30\,\mathrm{m}$ resolution and spatially heterogeneous precipitation on a $\sim 4\,\mathrm{km}$ grid. Among these $114$ storms, the April 17--21, 2013 flood-of-record corresponds to a physically consistent triplet of precipitation, basin topography, and observed gauge stages from the same storm; we use it to validate the underlying SynxFlow simulator against USGS water-surface-elevation observations at six basin gauges (four mainstem Des~Plaines and two tributary). The validation uses a per-gauge mean-recentered WSE series so that Nash-Sutcliffe Efficiency (NSE) reports timing-and-shape agreement after removing a systematic datum/bed-elevation offset; on this metric all six gauges have NSE in the $0.57$--$0.94$ range with an average of $0.81$ (Section~\ref{sec:illinois_validation}), establishing the simulator as a credible shape-level reference target for the surrogate. The dataset and trained checkpoints will be released publicly (Section~\ref{sec:conclusion}) to serve as a community benchmark for metropolitan-scale flood surrogates.
    \item[\textbf{(C4)}] \emph{Comprehensive multi-metric evaluation against three baselines.} We evaluate CLDNet against the unconditional LDNet (ablation of terrain conditioning), a VAE--ConvLSTM latent-dynamics baseline, and a Fourier Neural Operator along complementary axes (Section~\ref{sec:results}): aggregate rRMSE over valid evaluation points, water-depth time-series fidelity at high-impact flooded sites, flood-extent metrics (CSI, F1), and wall-clock efficiency. CLDNet improves the reported accuracy and extent metrics on both datasets while retaining second-scale inference at the full Illinois resolution.
    \item[\textbf{(C5)}] \emph{Assimilation-compatible latent state and meshless query interface.} The CLDNet latent state provides a compact dynamical representation of basin evolution, while the coordinate-based decoder can be queried at arbitrary locations, including the exact $(\mathrm{lon},\mathrm{lat})$ of USGS gauges (Section~\ref{sec:hydrograph_results}), without raster snapping. These demonstrated properties make the architecture structurally compatible with established latent ensemble-score-filter and differentiable-variational data-assimilation methods~\cite{si2025latentensf, xiao2026ldensf, levda2026}, with the validation pipeline of Section~\ref{sec:illinois_validation} supplying the natural observation locations at the six basin gauges. We therefore position CLDNet as a fast hydrodynamic surrogate layer toward AI-driven flood digital twins; online observation assimilation and time-varying antecedent state are immediate next experimental steps rather than demonstrated components of the present study.
\end{enumerate}

\subsection{Scale context and outline}

Table~\ref{tab:flood_problem_sizes} places the present work in context: the Illinois dataset has $4{,}188{,}840$ active watershed cells, roughly $6.9\times$ larger than the largest previously reported active- or area-equivalent flood-surrogate domain (ComGNN's Greens Bayou watershed, approximately $610{,}000$ cells at $30\,\mathrm{m}$ resolution). The Texas dataset ($250{,}000$ cells) is sized comparably to Hydro-Y-Net's Nan'gang River Basin test bed and serves as a controlled benchmark for comparing different surrogate-model families.

\begin{table}[!htb]
\centering
\small
\setlength{\tabcolsep}{6pt}
\renewcommand{\arraystretch}{1.15}
\begin{tabular}{lcc}
\toprule
\textbf{Method} & \textbf{Largest reported spatial size} & \textbf{Forcing} \\
\midrule
SWE-GNN~\cite{bentivoglio2023rapid}      & $16{,}384$ cells ($128\times 128$)      & Boundary inflow (dike breach) \\
mSWE-GNN~\cite{bentivoglio2025multi}      & $22{,}881$ cells       & Boundary inflow/outflow (dike breach) \\
U-FLOOD~\cite{lowe2021uflood}            & $65{,}536$ patch pixels ($256\times 256$)      & Uniform event hyetograph \\
Li et al.~\cite{li2023physical}          & $93{,}123$ space--time reference points  & Initial/boundary conditions \\
Hydro-Y-Net~\cite{yang2024rapid}         & $250{,}000$ grid cells     & Spatiotemporal rainfall \\
ComGNN~\cite{kazadi2024pluvial}          & $610{,}000$ area-equivalent cells  & Spatiotemporal rainfall \\
\midrule
Texas benchmark (this work)              & $250{,}000$ cells ($500\times 500$)     & Scalar hyetograph \\
Illinois / Des~Plaines (this work)       & $\mathbf{4{,}188{,}840}$ active cells & Spatiotemporal rainfall \\
\bottomrule
\end{tabular}
\caption{Largest active/computational cell counts or closest reported spatial extents in representative prior flood-surrogate studies and in the datasets introduced in this work. SWE-GNN and mSWE-GNN report computational cells; U-FLOOD reports $256\times256$ neural-network patches rather than active pixels; Li et al. report space--time reference points rather than active spatial cells; Hydro-Y-Net reports a $250{,}000$-grid forecasting region. For ComGNN, the largest Harris County region reported by Kazadi et al. is Greens Bayou: its watershed area is $549\,\mathrm{km}^2$, corresponding to roughly $610{,}000$ physical $30\,\mathrm{m}$ cells, while its bounding raster is $1{,}512\times 1{,}032$ pixels. The Illinois / Des~Plaines watershed contains $4{,}188{,}840$ active cells embedded in a $5{,}075\times 1{,}661$ raster, making the active watershed domain roughly $6.9\times$ larger than ComGNN's area-equivalent watershed cell count; furthermore, the $96$-hour prediction horizon represents a $28.8\times$ extension over the $200$-minute windows in ComGNN.}
\label{tab:flood_problem_sizes}
\end{table}

The remainder of the paper is organized as follows. Section~\ref{sec:dataset} states the governing SWEs, describes the SynxFlow simulator, and introduces the Texas benchmark and the Des~Plaines / Illinois case study with its USGS WSE validation. Section~\ref{sec:methodology} presents the CLDNet architecture, the terrain-conditioning mechanism, and the scalable training recipe. Section~\ref{sec:results} evaluates CLDNet against LDNet, VAE--ConvLSTM, and FNO baselines along the axes above. Section~\ref{sec:conclusion} summarizes the contributions and discusses dataset release and future directions toward operational and probabilistic flood forecasting.

\section{Problem setup and dataset generation}
\label{sec:dataset}

This section describes the high-fidelity flood datasets used to train and evaluate the proposed surrogate. We first state the governing shallow water equations (Section~\ref{sec:swe}) and the numerical solver used to generate reference trajectories (Section~\ref{sec:synxflow}). We then introduce two datasets of increasing realism: a synthetic Texas benchmark designed to isolate the effect of terrain conditioning under controlled rainfall variability (Section~\ref{sec:texas}) and a metropolitan-scale Illinois / Des~Plaines case study driven by real-rainfall stress-test forcings, whose simulator we validate against USGS gauge observations for the physically observed 2013 flood-of-record (Section~\ref{sec:illinois}). A consolidated summary is given in Section~\ref{sec:dataset_summary}.

\subsection{Governing equations}
\label{sec:swe}

Numerical hydrodynamic models simulate surface-water dynamics by solving the two-dimensional shallow water equations (SWEs) in conservative form,
\begin{equation}\label{eq:swe_original}
    \partial_t q + \div {F}(q) = R + S_b + S_f, \qquad (t,(x,y))\in [0, T]\times \Omega,
\end{equation}
with conserved variables, fluxes, and source terms
\begin{align*}
    q=
    \begin{bmatrix}
        h \\
        hu \\
        hv
    \end{bmatrix},\qquad
    F=\begin{bmatrix}
        uh & vh \\
        u^2h + \tfrac{1}{2}gh^2 & uvh \\
        uvh & v^2h + \tfrac{1}{2}gh^2
    \end{bmatrix},
\end{align*}
\begin{align*}
    R=
    \begin{bmatrix}
        r \\ 0 \\ 0
    \end{bmatrix},\qquad
    S_b=
    \begin{bmatrix}
        0 \\ -gh\partial_x b \\ -gh\partial_y b
    \end{bmatrix},\qquad
    S_f=
    \begin{bmatrix}
        0 \\ -C_f\, u \sqrt{u^2+v^2} \\ -C_f\, v \sqrt{u^2+v^2}
    \end{bmatrix},
\end{align*}
where $C_f = g n^2 h^{-1/3}$ is the Manning friction coefficient. The divergence $\div$ acts row-wise on $F$ with respect to $\xi = (x,y)$. We work throughout in the conservative variables $(h, hu, hv)$ (water depth and unit-width discharges), which are what the simulator outputs and what the surrogate in Section~\ref{sec:methodology} predicts; primitive velocities $u, v$ are recovered by $u=hu/h$ and $v=hv/h$ in wet cells. Because the friction coefficient $C_f$ is singular as $h\rightarrow 0$ and the velocity recovery $u=hu/h$ is ill-conditioned in shallow cells, SynxFlow regularizes the wet/dry treatment by using a small-depth threshold $h_{\min}=10^{-10}\,\mathrm{m}$: in the friction operator, cells with $h\le h_{\min}$ are assigned $u=0$ so the friction source term~$S_f$ vanishes; in the HLLC/augmented-Riemann interface fluxes, faces with both side depths $\le h_{\min}$ produce zero flux, and a one-sided dry-front Riemann state is used when only one side is dry. The implicit friction update of~\cite{xia2018new} additionally clips the friction force at the magnitude that would exactly cancel the cell's discharge over the time step, preventing momentum sign-flips in marginally wet cells. Notation is summarized in Table~\ref{tab:swe_notation}. The two datasets differ in the forcing $r(t,\xi)$, the terrain $b(\xi)$ and Manning coefficient $n(\xi)$, and the computational domain $\Omega$, while sharing the same solver.

\begin{table}[!htb]
\centering
\begin{tabular}{ll}
\toprule
\textbf{Symbol} & \textbf{Description} \\
\midrule
$h$ & Water depth \\
$u,\,v$ & Depth-averaged velocities in the $x$- and $y$-directions \\
$hu,\,hv$ & Unit-width discharges in the $x$- and $y$-directions \\
$g$ & Gravitational acceleration ($9.81\,\mathrm{m/s^2}$) \\
$b$ & Bed elevation \\
$n$ & Manning roughness coefficient \\
$C_f = g n^2 h^{-1/3}$ & Friction coefficient \\
$r$ & Rainfall rate \\
$q, F$ & Conserved state and interface flux \\
$S_b,\,S_f$ & Bed-slope and friction source terms \\
\bottomrule
\end{tabular}
\caption{Notation for the shallow water equations~\eqref{eq:swe_original}.}
\label{tab:swe_notation}
\end{table}

\subsection{High-fidelity simulator: SynxFlow}
\label{sec:synxflow}

Reference trajectories are produced with the open-source \href{https://github.com/SynxFlow/SynxFlow}{SynxFlow} solver~\cite{xia2017efficient, xia2018new, xia2019full}, a first-order Godunov-type finite-volume method for~\eqref{eq:swe_original} on structured rectangular grids. The key ingredients are: the surface reconstruction method (SRM) for well-balanced bed-elevation treatment, the HLLC Riemann solver for interface fluxes, a minmod-limited reconstruction of bed gradients, an explicit discretization of the flux and bed-slope terms, and an implicit treatment of the stiff friction term~\cite{xia2018new}. The time step is chosen adaptively from the CFL condition on gravity-wave and advective speeds. A self-contained description of the scheme is reported in Algorithm \ref{alg:FV-SWE}.

\begin{algorithm}[!htb]
\caption{Godunov-type finite-volume method for the shallow water equations}
\label{alg:FV-SWE}
\begin{algorithmic}[1]
    \STATE Discretize the rectangular computational domain into $N_{\mathrm{cell}}$ non-overlapping uniform rectangular cells~$\{\Omega_i\}$.
    \STATE Integrate Equation~\eqref{eq:swe_original} over each cell and apply the divergence theorem to obtain the semi-discrete finite-volume form
    \[
    \frac{\mathrm d q_i}{\mathrm d t}
    + \frac{1}{|\Omega_i|}\sum_{k\in\mathcal N_i} F_k(q)\, l_k
    = R_i + S_{b,i} + S_{f,i},
    \]
    where $q_i$ denotes the cell-averaged conserved variables, $\mathcal N_i$ is the set of faces of cell $\Omega_i$, and $l_k$ is the length of face $k$.
    \STATE At time level $n$, reconstruct the bed elevation and water-surface states at each cell interface using the surface reconstruction method (SRM); compute the limited bed gradients with the minmod limiter.
    \STATE Construct the left and right Riemann states at each interface from the reconstructed quantities.
    \STATE Evaluate the numerical fluxes $F_k^n$ at all cell interfaces using the HLLC Riemann solver.
    \STATE Compute the bed-slope source term $S_{b,i}^n$ from the reconstructed interface states.
    \STATE Choose the time step from the CFL condition
    \[
        \Delta t = \mathrm{CFL}\min_i
        \frac{d_i}{\sqrt{u_i^2+v_i^2}+\sqrt{g h_i}},
        \qquad 0<\mathrm{CFL}\le 1,
    \]
    where $d_i$ is the minimum distance from the cell center to its edges.
    \STATE Update the conserved variables by
    \[
        q_i^{n+1}
        = q_i^n
        - \frac{\Delta t}{|\Omega_i|}\sum_{k\in\mathcal N_i} F_k^n l_k
        + \Delta t\left(R_i^n + S_{b,i}^n + S_{f,i}^{\,n+1}\right),
    \]
    where the flux and bed-slope terms are discretized explicitly, while the friction contribution $S_{f,i}^{\,n+1}$ is evaluated implicitly using the scheme of~\cite{xia2018new}.
    \STATE Impose the prescribed physical boundary conditions through ghost-cell or interface states.
    \STATE Advance to the next time step until the final simulation time is reached.
\end{algorithmic}
\end{algorithm}

\paragraph{Hardware and runtime.}
All SynxFlow simulations are performed on a single NVIDIA L40S GPU and AMD EPYC 9334 CPU (unless stated otherwise in Section~\ref{sec:training}). The average wall-clock time per (96-hour) Illinois trajectory is approximately $55$ minutes (see Section~\ref{sec:illinois}) whereas each (48-hour) Texas trajectory requires about $2$ minutes (see Section~\ref{sec:texas}).

\subsection{Synthetic Texas benchmark}
\label{sec:texas}

The Texas dataset is a controlled benchmark for evaluating rainfall-driven flood surrogates on a fixed but nontrivial real-terrain domain, isolating the effect of terrain conditioning under rainfall variability while avoiding the confounding factors of real-world forcing heterogeneity and channel-network complexity. It uses a single real-terrain DEM with spatially uniform, time-varying rainfall over a $15\,\mathrm{km}\times 15\,\mathrm{km}$ domain ($500\times 500 = 250{,}000$ cells). The grid size is chosen near the upper limit at which the regular-grid baselines remain trainable on a single high-end GPU, so that all models can be compared on the same domain.

\paragraph{Domain and terrain.}
The DEM used for this dataset is derived from the USGS 3D Elevation Program (3DEP). The center of the sampled domain is located in Bell County, central Texas, with coordinates at approximately 30.9883 degrees North, 97.7274 degrees West. It represents a square domain at $30\,\mathrm{m}$ resolution, giving $500\times 500=250{,}000$ grid cells (see Figure~\ref{fig:texas_illinois_dem}\subref{fig:texas}). The elevation spans $190\,\mathrm{m} -334\,\mathrm{m}$ (mean: $245\,\mathrm{m}$) with a mean slope of 0.04, yielding terrain heterogeneity sufficient to produce nontrivial flow patterns under spatially uniform forcing. All cells are assigned a single Manning coefficient $n=0.035$, and open boundary conditions are imposed so that water exiting the domain is removed without reflection.

\begin{figure}[!htb]
\centering
\includegraphics[width=0.95\linewidth]{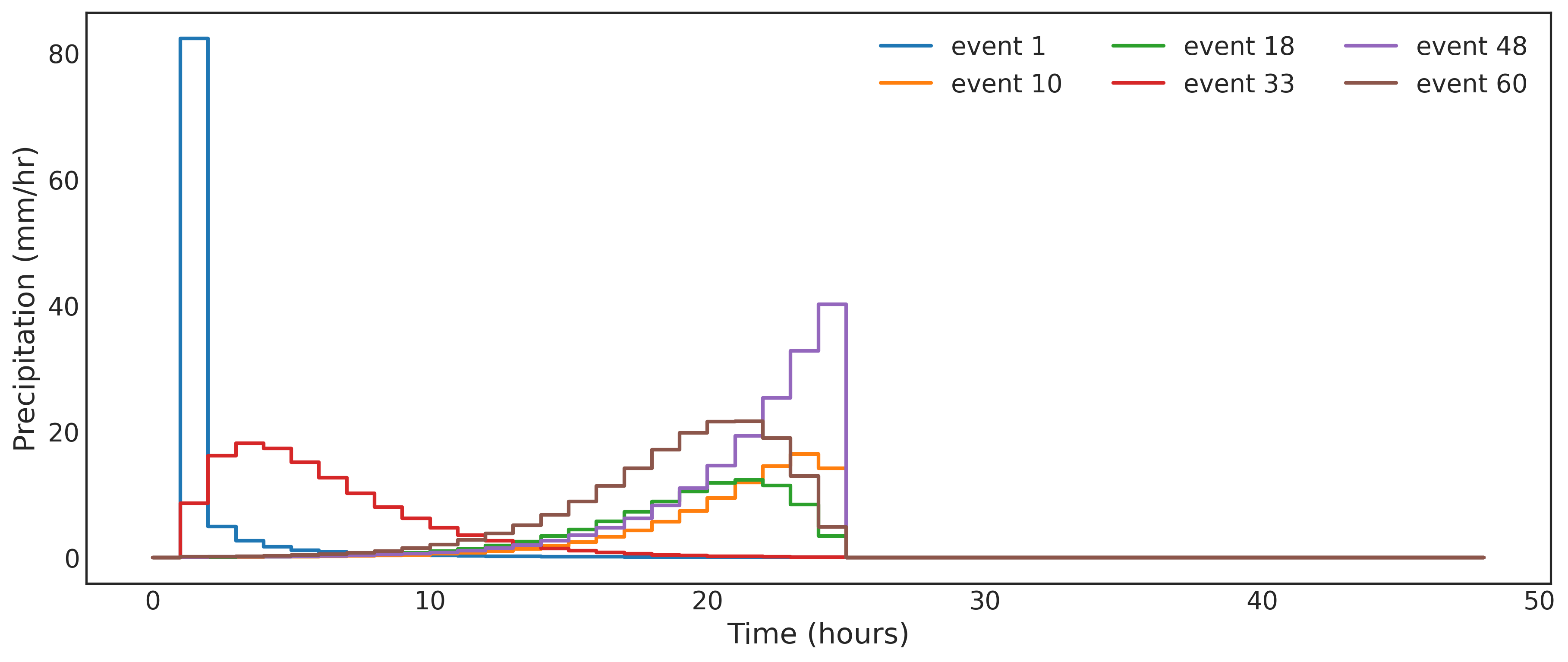}
\caption{Representative spatially uniform precipitation hyetographs for the Texas benchmark. Six randomly selected events demonstrate the temporal rainfall variability and intensity ranges within the dataset.}
\label{fig:texas_precip_examples}
\end{figure}

\paragraph{Precipitation forcing.}
A spatially uniform precipitation rate $r(t)$ is prescribed. A total of $120$ hyetographs are generated, of which half are time-reversed copies of the other half; this construction exposes the surrogate to both rising-limb-dominated and recession-dominated rainfall shapes. Each of the $120$ hyetographs is simulated over the terrain by SynxFlow independently and yields a distinct flood trajectory. The $100$-train / $20$-test split is over these trajectories. Six examples are shown in Figure~\ref{fig:texas_precip_examples}.

\paragraph{Simulation output.}
Each trajectory is integrated over $T=48\,\mathrm{h}$ with output snapshots saved every hour, giving $N_T=49$ snapshots of $(h, hu, hv)$ on the full $500\times 500$ grid.

\subsection{Des Plaines River basin case study, Illinois}
\label{sec:illinois}

The Illinois dataset anchors the paper in a realistic, operationally relevant setting. It targets the Des~Plaines River basin in northeastern Illinois, an eight-digit Hydrologic Unit Code (HUC8~07120004) watershed that drains the greater Chicago metropolitan area, and is driven by real Stage~IV rainfall fields used as stress-test forcings. Two aspects distinguish it from prior flood surrogate benchmarks. First, the computational domain contains $4{,}188{,}840$ active watershed cells at $30\,\mathrm{m}$ resolution, roughly $6.9\times$ larger than the largest prior active- or area-equivalent domain in Table~\ref{tab:flood_problem_sizes}. Second, for the physically observed 2013 flood-of-record, we validate the underlying SynxFlow simulator against USGS gauge observations, establishing that the surrogate's reference data are credible at event scale for that event.

\subsubsection{Domain and static inputs}
\label{sec:illinois_domain}

\begin{figure}[!htb]
\centering
\begin{subfigure}{0.5\linewidth}
    \centering
    \includegraphics[width=\linewidth]{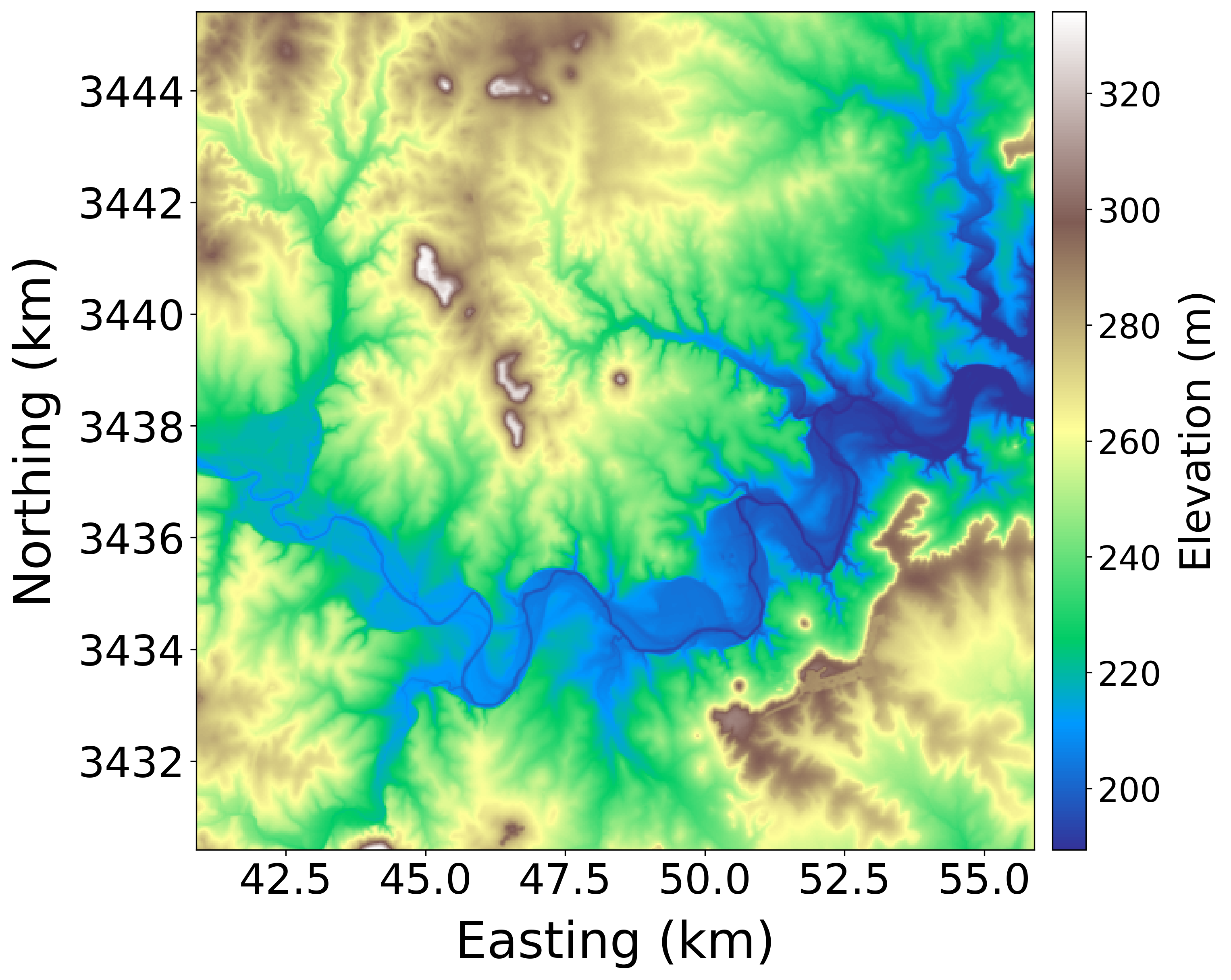}
    \caption{Texas DEM}
    \label{fig:texas}
\end{subfigure}
\hspace{2em}
\begin{subfigure}{0.3\linewidth}
    \centering
    \includegraphics[width=\linewidth]{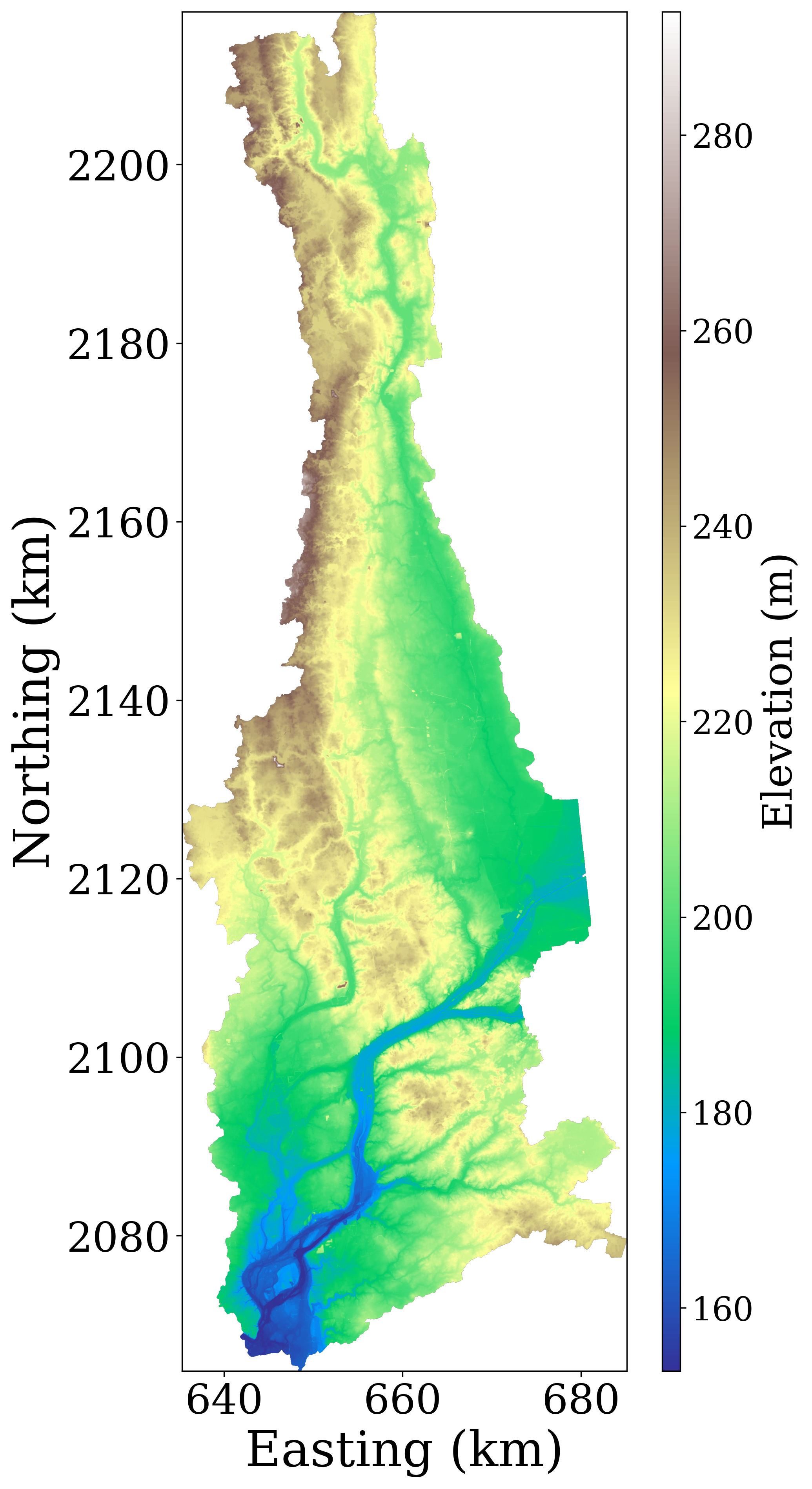}
    \caption{Des~Plaines / Illinois DEM}
    \label{fig:illinois}
\end{subfigure}
\caption{Digital elevation models used in the Texas benchmark and the Des Plaines / Illinois case study at $30\,\mathrm{m}$ resolution. (a) Square $15\,\mathrm{km} \times 15\,\mathrm{km}$ domain in Bell County, Texas ($500 \times 500 = 250{,}000$ cells). (b) Des Plaines River basin (HUC8~07120004) featuring $4{,}188{,}840$ active watershed cells derived from USGS 3DEP.}
\label{fig:texas_illinois_dem}
\end{figure}

The Illinois dataset uses a non-rectangular active computational domain with $4{,}188{,}840$ watershed cells embedded in a $5,075 \times 1,661$ raster at $30\,\mathrm{m}$ resolution. Cells outside the basin are excluded from the active domain and masked with \texttt{NaN} (Figure~\ref{fig:texas_illinois_dem}\subref{fig:illinois}). The DEM is derived from USGS 3DEP, and the drainage area of the basin is approximately $3,814\,\mathrm{km}^2$. Two Manning coefficients are used to encode first-order land-cover heterogeneity: $n=0.02$ for water-covered cells and $n=0.05$ for land-covered cells, classified from NLCD year of 2021. Open boundary conditions are imposed at the downstream outlet and along the domain edge.

\subsubsection{Precipitation forcing and storm events}
\label{sec:illinois_forcing}

Precipitation is spatially heterogeneous and is provided on a coarser $39\times 13$ grid of about $4\,\mathrm{km}\times 4\,\mathrm{km}$ pixels, resampled onto the computational grid during simulation. The dataset contains $114$ storm events drawn from NCEP Stage~IV Quantitative Precipitation Estimates (QPE) between 2002 and 2024.

The $114$ events are not all storms that historically struck the Des~Plaines basin. Instead, we treat the Stage~IV archive as a \emph{library} of high-intensity rainfall fields that we apply as forcing on the Des~Plaines DEM, in order to obtain a diverse set of stress-test trajectories well beyond what the basin's own 22-year record would yield. Sampling proceeds in two steps. First, candidate centroids are drawn uniformly at random within each of twelve Midwestern states (Illinois, Indiana, Iowa, Kansas, Michigan, Minnesota, Missouri, Nebraska, North Dakota, Ohio, South Dakota, and Wisconsin), and the candidates with the highest cumulative 2002--2024 rainfall are retained. Second, for each retained centroid, we extracted the top ten events based on peak rainfall intensity. This yielded a total of 114 events, noting that six additional events were unavailable for download. The April 17--21, 2013 flood-of-record (Section~\ref{sec:illinois_validation}) provides a physically consistent observation cross-check, in which the rainfall, topography, and observed gauge stages all correspond to the same physical storm and the rainfall field is a Stage~IV product covering the actual Des~Plaines basin during that storm.

\begin{figure}[!htb]
\centering
\includegraphics[width=0.95\linewidth]{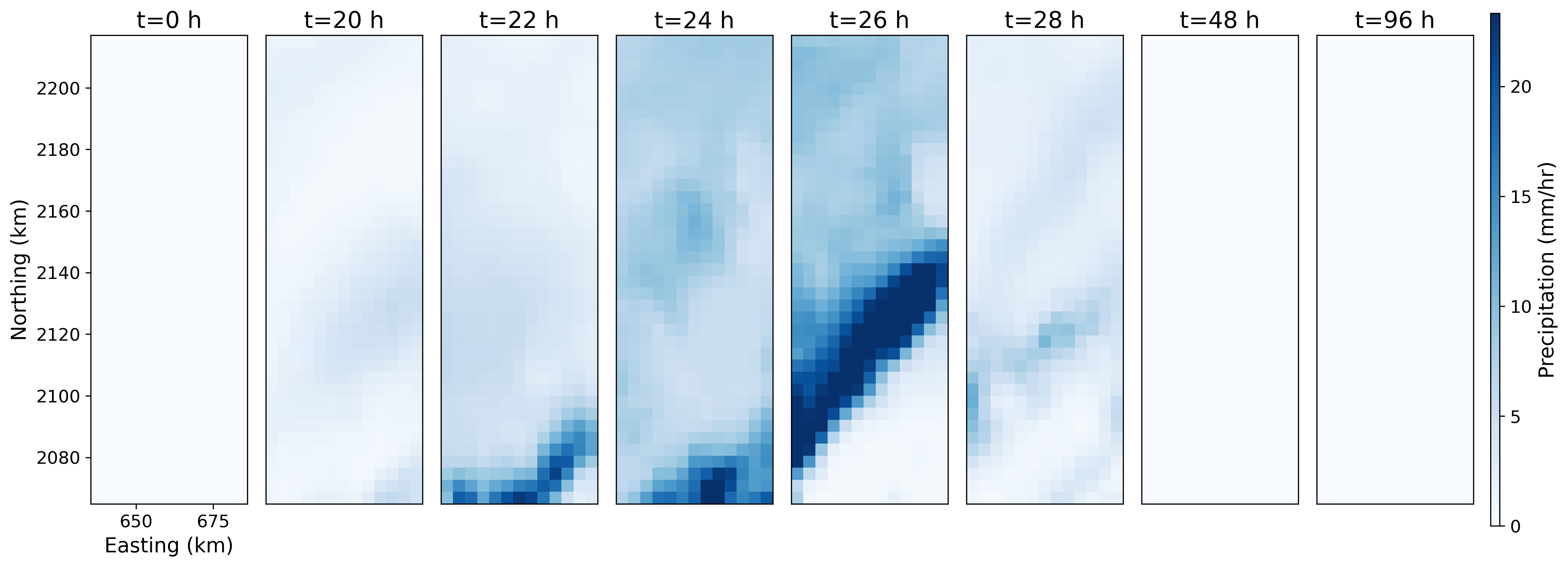}
\caption{Two-dimensional precipitation field for one representative storm event at $t=0,20,22,24,26,28,48,96\,\mathrm{h}$.}
\label{fig:illinois_precip}
\end{figure}

\begin{figure}[!htb]
\centering
\includegraphics[width=0.95\linewidth]{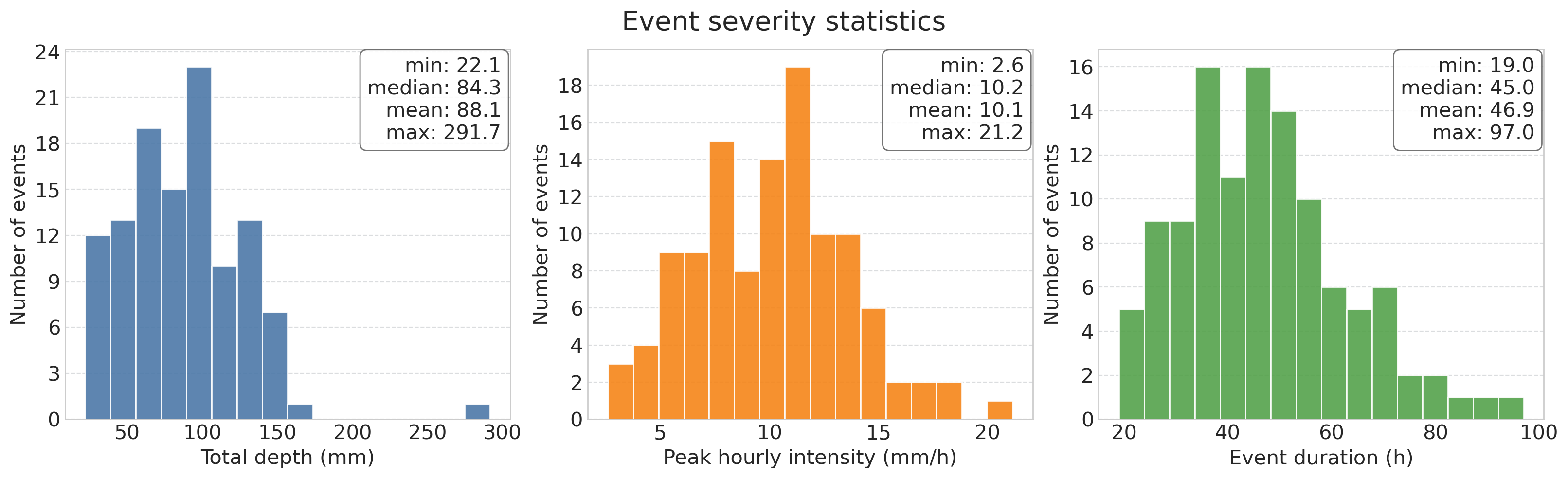}
\caption{Storm-event severity distribution and summary statistics across the 114 Illinois events.}
\label{fig:illinois_event_stats}
\end{figure}

Figure~\ref{fig:illinois_precip} shows the two-dimensional precipitation field for a representative event at selected times. Event-severity statistics (total depth, peak hourly intensity, event duration) are summarized in Figure~\ref{fig:illinois_event_stats}.

\subsubsection{Simulation protocol and computational cost}
\label{sec:illinois_protocol}

To avoid a cold-start bias across all $114$ events, we use a two-stage spin-up procedure. The long 2013-04-10 to 2013-04-23 storm is first simulated from a dry-bed initial condition. The resulting flow state at $t=175\,\mathrm{h}$ is used as the shared initial condition for all $114$ forecast windows, including the held-out 2013 event and the $113$ stress-test events. For each event, we extract the window containing the bulk of the precipitation and integrate the system for $96\,\mathrm{h}$ with an output interval of $1\,\mathrm{h}$, yielding $N_T=97$ snapshots per trajectory. Each snapshot stores $(h, hu, hv)$ on the in-domain cells, giving a tensor of shape $(97, 3, 5075, 1661)$ per trajectory. The $114$ trajectories are split into $100$ training, $10$ validation, and $4$ test trajectories. The 2013-04-17 to 2013-04-21 flood-of-record event used for USGS WSE validation in Section~\ref{sec:illinois_validation} is one of the four held-out test trajectories; it never appears in the training or validation splits, so all surrogate-level metrics reported on it (Table~\ref{tab:traj_rmse}) are evaluated on a strictly held-out event. The remaining three test trajectories are reported in aggregate as ``Non-2013 held-out test average'' and are similarly excluded from training and validation. The average wall-clock cost per trajectory is about $55$ minutes on the hardware reported in Section~\ref{sec:synxflow}, for a cumulative simulation budget of $105$ GPU hours across all events.

\subsubsection{Validation of the simulator against USGS observations}
\label{sec:illinois_validation}

To establish that the SynxFlow outputs are a credible training target for the surrogate, we validate the simulator against USGS stream-gauge observations during the 2013-04-17 to 2013-04-21 Des~Plaines flood event, one of the most recent destructive events in the basin.

\begin{figure}[!htb]
\centering
\includegraphics[width=\linewidth]{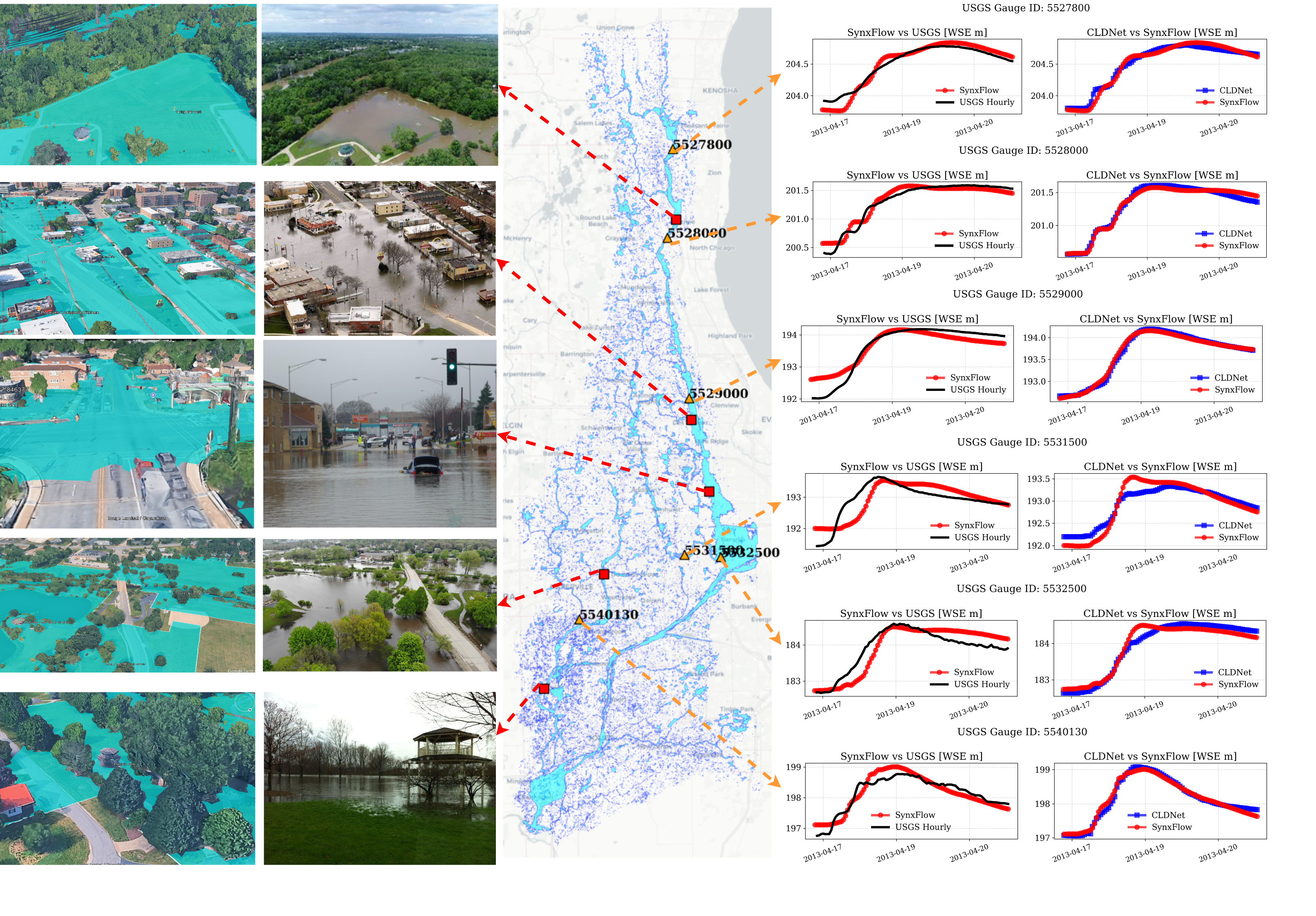}
\caption{Validation of the SynxFlow simulator against USGS WSE observations for the 2013-04-17 to 2013-04-21 Des~Plaines flood event. Left: aerial photographs of flooded reaches at the six gauge locations. Center: watershed map with NHDPlus flowlines and USGS gauge markers; red dashed lines link photographs to gauge locations. Right: WSE time series at each gauge from SynxFlow (red), USGS (black), and CLDNet (blue) for visual reference; full surrogate evaluation in Section~\ref{sec:results}.}
\label{fig:desplaines_validation}
\end{figure}

\paragraph{Gauge set.}
We use six USGS stream gauges in the Des~Plaines basin for simulator validation (Figure~\ref{fig:desplaines_validation}): four mainstem Des~Plaines River gauges (05527800 At~Russell, 05528000 Near~Gurnee, 05529000 Near~Des~Plaines, 05532500 At~Riverside) and two tributary gauges (05531500 Salt~Creek At~Western~Springs, 05540130 West~Branch~Du~Page~River Near~Naperville). These gauges are selected because their channels are sufficiently resolved at $30\,\mathrm{m}$ DEM resolution that local hydraulic controls (weirs, culverts, urban channels) do not dominate the comparison.

\paragraph{Cross-section construction.}
At each gauge, we generate a perpendicular cross-section along the associated NHDPlus flowline with a width of $60\,\mathrm{m}$ (extended to $90\,\mathrm{m}$ for four wider-channel gauges) and map the cross-section onto the simulation raster. For validation we compare water-surface elevation (WSE), computed from the SynxFlow water-depth and bed-elevation fields at the raster cells intersected by each cross-section.

\paragraph{Reference data.}
Stage data are retrieved from the USGS National Water Information System (NWIS) using the pygeohydro python library \cite{Chegini_2021}. We use the Instantaneous Value (IV) web service to retrieve high-resolution sub-hourly stage observations and convert them to WSE for comparison against SynxFlow.

\paragraph{Visual comparison.}
Figure~\ref{fig:desplaines_validation} overlays SynxFlow, CLDNet, and USGS WSE time series for the six gauges, alongside aerial photographs of flooded reaches at the corresponding locations during the 2013 event. The CLDNet curve is included for context only; the validation claim in this section concerns the SynxFlow reference simulator. To make the WSE-shape agreement visible despite a residual datum/bed-elevation offset between the simulator and the gauge, the simulated WSE has been recentered on the observed mean: at each gauge, we subtract the simulator's mean WSE over the comparison window and add the observed (USGS) mean WSE over the same window, so that the shifted simulated series and the observed series share the same temporal mean by construction. The metrics in Table~\ref{tab:desplaines_validation_metrics} below are computed on this recentered series. After this correction, SynxFlow tracks the temporal shape and peak timing of both the rising limb and the peak closely; residual discrepancies are most visible in the recession phase at the two tributary gauges, where local hydraulic controls not represented in the DEM (small weirs, culverts) modulate the observed stage.

\paragraph{Quantitative WSE metrics.}
For the simulator validation, we report Nash--Sutcliffe efficiency (NSE) on a per-gauge mean-recentered simulated WSE series. The recentering removes a systematic datum/bed-elevation offset between the simulated and observed water surfaces. The simulator's per-cross-section WSE is constructed as $\mathrm{WSE}_{\mathrm{sim}} = \max_{\xi\in\mathrm{CS}}h(\xi) + \overline{b}_{\mathrm{CS}}$, i.e., the maximum simulated water depth across the cross-section pixels plus the mean DEM elevation along the same cross-section, while the observed WSE is the USGS gauge stage plus the gauge's published altitude datum, $\mathrm{WSE}_{\mathrm{obs}} = \mathrm{stage}_{\mathrm{USGS}} + \mathrm{alt\_datum}$. Differences between $\overline{b}_{\mathrm{CS}}$ (a $30\,\mathrm{m}$ DEM average) and the gauge's surveyed reference datum produce a per-gauge constant offset that is removed by the recentering described above, so what we report is timing-and-shape agreement rather than agreement on raw absolute WSE. Table~\ref{tab:desplaines_validation_metrics} summarizes the six gauges. All six show consistently positive shape-level WSE skill, with NSE ranging from $0.566$ to $0.944$, with an average of $0.815$. We do not report raw WSE NSE because it is dominated by the constant offset and is not a meaningful measure of dynamic agreement at this scale.

\begin{table}[!htb]
\centering
\small
\setlength{\tabcolsep}{6pt}
\renewcommand{\arraystretch}{1.15}
\begin{tabular}{llc}
\toprule
\textbf{USGS ID} & \textbf{Station} & \textbf{NSE (WSE)} \\
\midrule
05527800 & Des Plaines River At Russell, IL & $0.910$ \\
05528000 & Des Plaines River Near Gurnee, IL & $0.944$ \\
05529000 & Des Plaines River Near Des Plaines, IL & $0.867$ \\
05531500 & Salt Creek At Western Springs, IL & $0.566$ \\
05532500 & Des Plaines River At Riverside, IL & $0.746$ \\
05540130 & West Branch Du Page River Near Naperville, IL & $0.854$ \\
\bottomrule
\end{tabular}
\caption{SynxFlow vs.\ USGS WSE validation at the six Des~Plaines basin gauges (four mainstem, two tributary) for the April 17--21, 2013 event. NSE is computed on a per-gauge mean-recentered simulated WSE series (simulated mean replaced with the observed mean over the comparison window) so that it reflects temporal-shape agreement after removing the systematic datum/bed-elevation offset; values are not directly comparable to literature NSE on raw WSE.}
\label{tab:desplaines_validation_metrics}
\end{table}

\paragraph{Implication for the surrogate.}
The surrogate developed in Section~\ref{sec:methodology} is trained against the SynxFlow reference; all surrogate-level error metrics reported in Section~\ref{sec:results} are therefore measured relative to a simulator that itself has been validated against independent USGS observations at event scale. Residual simulator biases (e.g., sub-grid hydraulic structures) propagate into both the training target and the test target, so surrogate errors near those gauges should be interpreted as errors with respect to SynxFlow rather than with respect to ground truth.

\subsection{Summary of datasets}
\label{sec:dataset_summary}

Table~\ref{tab:dataset_summary} consolidates the properties of the two datasets for reference.

\begin{table}[!htb]
\centering
\small
\setlength{\tabcolsep}{6pt}
\renewcommand{\arraystretch}{1.15}
\begin{tabular}{lll}
\toprule
& \textbf{Texas (synthetic)} & \textbf{Illinois / Des~Plaines (real-rainfall)} \\
\midrule
Domain shape                & Square                            & Non-rectangular (\texttt{NaN}-masked) \\
Grid size                   & $500\times 500$                   & $5{,}075\times 1{,}661$ \\
Active in-domain cells      & $250{,}000$                       & $4{,}188{,}840$ \\
Spatial resolution          & $30\,\mathrm{m}$                  & $30\,\mathrm{m}$ \\
DEM source                  & USGS 3DEP                         &  USGS 3DEP \\
Manning coefficient         & Uniform $0.035$                   & $0.02$ (water) / $0.05$ (land) \\
Precipitation forcing       & Scalar, spatially uniform         & $39\times 13$ field, $\sim 4\,\mathrm{km}$ \\
Precipitation source        & Synthetic hyetographs             & Stage~IV QPE, 2002--2024 \\
\# trajectories             & $120$                               & $114$ \\
Simulation horizon          & $48\,\mathrm{h}$                  & $96\,\mathrm{h}$ \\
Output time interval        & $1\,\mathrm{h}$                & $1\,\mathrm{h}$ \\
Snapshots per trajectory    & $N_T=49$                         & $N_T=97$ \\
Boundary condition         & Open                              & Open \\
Wall-clock time/trajectory   & $2\,\mathrm{min}$                 & $55\,\mathrm{min}$ \\
Simulator validation        & N/A (synthetic)                   & 6 USGS gauges (4 mainstem, 2 tributary), 2013 event \\
\bottomrule
\end{tabular}
\caption{Comparative overview of dataset characteristics: the synthetic Texas benchmark (spatially uniform forcing on a square domain) and the Des Plaines, Illinois case study (real-rainfall stress-test forcings on a $30\,\mathrm{m}$-resolution, complex basin geometry).}
\label{tab:dataset_summary}
\end{table}

\section{Methodology}
\label{sec:methodology}

\subsection{Problem statement}
\label{sec:problem}

We seek a surrogate for the shallow-water system of Section~\ref{sec:dataset} that maps rainfall forcing and static terrain descriptors to the space--time evolution of the conserved flow state. Let $\Omega\subset\mathbb{R}^2$ denote the physical domain and let $[0,T]$ be the simulation horizon discretized into $N_T$ uniform snapshots $\{t_0,t_1,\dots,t_{N_T-1}\}$ with step $\Delta t$. At each spatial location $\xi\in\Omega$ and time $t_k$, let
\begin{equation}
    q(t_k,\xi) \;=\; \big(h(t_k,\xi),\, hu(t_k,\xi),\, hv(t_k,\xi)\big)\in\mathbb{R}^3
    \label{eq:state}
\end{equation}
denote the conserved variables (water depth and unit-width discharges), matching the notation of Section~\ref{sec:swe}, and let $r_{t_k}$ denote the rainfall forcing at time $t_k$. The forcing is a scalar for the Texas dataset (spatially uniform precipitation) and a $39\times 13$ coarse-grid field for the Illinois dataset (spatially heterogeneous precipitation); in both cases it is resampled onto a shared temporal grid with the simulation snapshots. Static terrain information is summarized by the bed elevation $b(\xi)$ and the Manning roughness $n(\xi)$, which are available on the fine simulation grid.

Given a set of training trajectories
\begin{equation}
    \mathcal{D}_{\mathrm{train}} \;=\; \Big\{\big(r^{(i)}_{0:N_T-1},\,q^{(i)}(t_{0:N_T-1},\,\Omega_h)\big)\Big\}_{i=1}^{N_{\mathrm{traj}}},
\end{equation}
where $\Omega_h\subset\Omega$ is the discrete simulation grid, our goal is to learn a surrogate map
\begin{equation}
    \mathcal{S}_\theta\colon\; \big(r_{0:N_T-1},\, b,\, n\big) \;\longmapsto\; \big\{\tilde{q}(t_k,\xi)\big\}_{k=0,\dots,N_T-1,\,\xi\in\Omega_{\mathrm{query}}}
    \label{eq:surrogate}
\end{equation}
that, at inference time, can be queried at an arbitrary set of spatial points $\Omega_{\mathrm{query}}\subseteq\Omega$ and is substantially cheaper to evaluate than the high-fidelity simulator. The map~\eqref{eq:surrogate} assumes a fixed basin initial condition: on Texas this is a dry bed, and on Illinois this is the shared post-spin-up state described in Section~\ref{sec:illinois_protocol}. Conditioning on a varying antecedent flow state $q_0$ to support, e.g., operational forecasting from current basin conditions is a natural extension that we leave to future work (Section~\ref{sec:limitations}). The surrogate should (i) respect terrain heterogeneity that controls local flood dynamics, (ii) scale to metropolitan-size grids with millions of cells, and (iii) support fast, memory-efficient inference for large ensemble and scenario-screening workflows. The remainder of this section develops a conditional latent-dynamics architecture that meets these requirements.

\subsection{Latent Dynamics Networks}
\label{sec:ldnet}

Latent Dynamics Networks (LDNet), originally introduced by Regazzoni et al.~\cite{regazzoni2024learning} and subsequently refined by Xiao et al.~\cite{xiao2026ldensf} in the context of ensemble-score-filter data assimilation, provide a compact, meshless surrogate paradigm that decouples temporal evolution from spatial decoding. Rather than propagating the full field forward in time, an LDNet advances a low-dimensional latent state $s_t\in\mathbb{R}^{d_s}$ through a learned neural dynamical system and reconstructs the physical field on demand at arbitrary query coordinates through a coordinate-based decoder. This factorization yields two structural advantages that are particularly attractive for regional flood modeling: the memory cost of a forward pass scales with the number of query points rather than the full grid size, and the decoder is naturally positioned to accept additional spatial conditioning signals (Section~\ref{sec:cldnet}). The same latent-state paradigm has subsequently been extended to high-dimensional data assimilation~\cite{si2025latentensf, xiao2026ldensf, levda2026} and to coordinate-based reconstruction from sparse sensors~\cite{stride2026}, establishing it as a general framework for spatiotemporal surrogate modeling under partial observations.

Concretely, the model consists of two neural networks: a dynamics network $F_{\theta_1}$ that estimates the temporal derivative of the latent state given the current latent state and the external forcing, and a reconstruction network $\mathcal{R}_{\theta_2}$ that maps a latent state and a query coordinate to the corresponding field value. Given a forcing sequence $\{r_{t_k}\}_{k=0}^{N_T-1}$, the latent state is initialized one step before the first snapshot, $s_{t_{-1}}=0$, and evolved by an explicit forward-Euler step of a neural ordinary differential equation (ODE)
\begin{equation}
    s_{t_k} \;=\; s_{t_{k-1}} + \Delta t\, F_{\theta_1}\!\big(s_{t_{k-1}},\, r_{t_k}\big),
    \qquad k=0,1,\dots,N_T-1.
    \label{eq:latent_euler}
\end{equation}
The shift from the $s_{t_0}=0$ initialization of the original formulation~\cite{regazzoni2024learning} to $s_{t_{-1}}=0$ follows the refinement in~\cite{xiao2026ldensf}, which observed that initializing the latent at $s_{t_0}=0$ prevents the first decoding at $t_0$ from depending on the forcing and, consequently, fails to capture problems with varying initial conditions or strong early-time response --- a regime that applies to flood events with early rising-limb dynamics. At any time $t_k$ and query coordinate $\xi\in\Omega$, the field is reconstructed as
\begin{equation}
    \tilde{q}(t_k,\xi) \;=\; \mathcal{R}_{\theta_2}\!\big(s_{t_k},\, \gamma(\xi)\big),
    \label{eq:vanilla_decoder}
\end{equation}
where $\gamma(\xi)$ is a Fourier feature mapping of the normalized coordinate that mitigates spectral bias of coordinate MLPs~\cite{tancik2020fourier, xiao2026ldensf}. Both $F_{\theta_1}$ and $\mathcal{R}_{\theta_2}$ are multilayer perceptrons. Architectural details are reported in Section~\ref{sec:results}.

\paragraph{Forcing representation.}
For the Texas dataset the forcing $r_{t_k}\in\mathbb{R}$ is the instantaneous scalar precipitation rate and is passed directly to the dynamics network. For the Illinois dataset the forcing is the $39\times 13$ precipitation field, which we flatten into a $507$-dimensional vector and concatenate with the latent state before passing to the dynamics network $F_{\theta_1}$.

\subsection{Conditional Latent Dynamics Networks for Flooding}
\label{sec:cldnet}

Local flood dynamics are strongly controlled by static terrain characteristics: elevation gradients set the direction and speed of overland flow, slope magnitude modulates momentum transfer, and Manning roughness controls bottom friction. Under a plain LDNet decoder~\eqref{eq:vanilla_decoder}, this information is available only implicitly through the latent state $s_{t_k}$ and the raw coordinate~$\xi$. Empirically (Section~\ref{sec:results}), this is insufficient at a regional scale: the decoder must effectively memorize a $4.2$M-active-cell terrain field through the coordinate input alone, which the latent state cannot disambiguate.

We therefore introduce an explicit static spatial conditioning vector $\phi(\xi)\in\mathbb{R}^{d_\phi}$ that is concatenated with the (Fourier-embedded) coordinate before being passed to the decoder. The conditional decoder reads
\begin{equation}
    \tilde{q}(t_k,\xi) \;=\; \mathcal{R}_{\theta_2}\!\Big(s_{t_k},\, \big[\gamma(\xi),\, \phi(\xi)\big]\Big),
    \qquad k=0,1,\dots,N_T-1,\ \ \xi\in\Omega,
    \label{eq:cond_decoder}
\end{equation}
where $[\,\cdot\,,\,\cdot\,]$ denotes concatenation along the feature dimension.
This preserves the latent evolution~\eqref{eq:latent_euler} unchanged while enriching the spatial input of the decoder with terrain-derived features. 

\paragraph{Static conditioning features.}
Our default choice is $\phi(\xi)=\big(b_{\mathrm{z}}(\xi),\, b_g(\xi),\, \tilde n(\xi)\big)\in\mathbb{R}^3$, consisting of a standardized elevation, the slope magnitude, and a scaled Manning coefficient. The standardized elevation is
\begin{equation}
    \mu_b \;=\; \frac{1}{|\Omega_h|}\sum_{\xi\in\Omega_h} b(\xi),
    \qquad
    \sigma_b \;=\; \sqrt{\frac{1}{|\Omega_h|}\sum_{\xi\in\Omega_h}\big(b(\xi)-\mu_b\big)^2},
    \qquad
    b_{\mathrm{z}}(\xi) \;=\; \frac{b(\xi)-\mu_b}{\sigma_b},
\end{equation}
which removes the large mean elevation offset that otherwise dominates the decoder input. The slope magnitude is
\begin{equation}
    b_g(\xi) \;=\; \big\|\nabla b(\xi)\big\|_2
    \;=\; \sqrt{\big(\partial_x b(\xi)\big)^2 + \big(\partial_y b(\xi)\big)^2},
\end{equation}
computed on the fine simulation grid via centered finite differences. The scaled Manning coefficient $\tilde n(\xi)=100\,n(\xi)$ simply rescales the small Manning values to order-one magnitudes so that they are comparable to the other entries of $\phi(\xi)$; on the Illinois dataset the two Manning values ($0.02$ for water-covered cells and $0.05$ for land cells) map to $\tilde n \in \{2, 5\}$. Section~\ref{sec:results} quantifies the effect of this conditioning by comparing CLDNet against the unconditional LDNet baseline on both datasets.

\begin{figure}[!htb]
  \centering
  \includegraphics[width=0.5\linewidth]{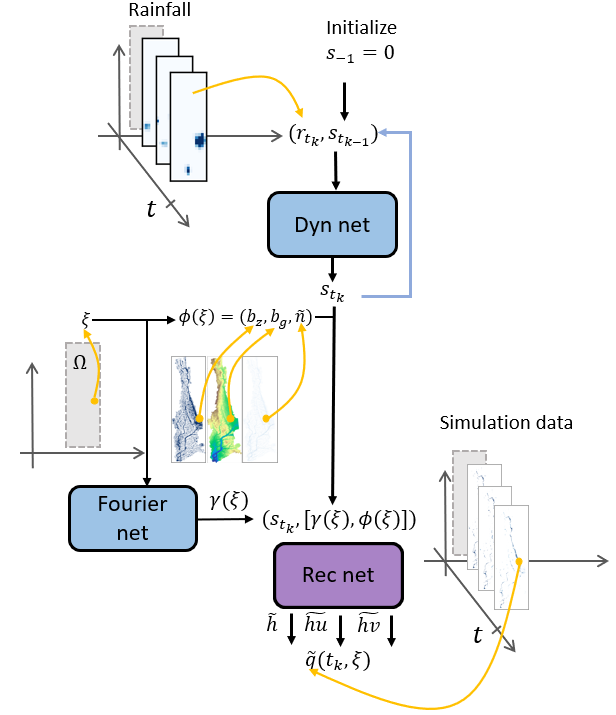}
  \caption{Conditional LDNet (CLDNet). Starting from an initial zero latent state, at each time step $t_k$, rainfall forcing $r_{t_k}$ and the previous latent state $s_{t_{k-1}}$ drive a latent neural ODE~\eqref{eq:latent_euler} via the dynamics network (Dyn net) to produce the updated latent state $s_{t_k}$. The reconstruction network (Rec net)~\eqref{eq:cond_decoder} then consumes the tuple $(s_{t_k}, [\gamma(\xi), \phi(\xi)])$ to decode the flow field of water depth $\tilde h$ and unit discharges $\widetilde{hu}, \widetilde{hv}$ at an arbitrary query coordinate $\xi \in \Omega$. The spatial conditioning relies on a Fourier feature embedding $\gamma(\xi)$ and a static terrain feature vector $\phi(\xi) = (b_z, b_g, \tilde n)$ derived from DEM and Manning data.}
    \label{fig:cldnet_arch}
\end{figure}

Figure~\ref{fig:cldnet_arch} summarizes the full CLDNet forward pass. The dynamics network rolls the latent state forward over the full trajectory in a single autoregressive pass; for each queried $(t_k,\xi)$ pair, the reconstruction network consumes the corresponding latent state, the Fourier-embedded coordinate, and the static conditioning vector to produce $\tilde q(t_k,\xi)$.

\subsection{Training strategy and scalability}
\label{sec:training}

The decoder~\eqref{eq:cond_decoder} is evaluated pointwise, so training cost and memory scale with the number of query points used for the loss rather than with the full simulation-grid size $|\Omega_h|$. We exploit this property to train at a metropolitan scale on a single compute node.

\paragraph{Loss.}
Let $\{\xi_m\}_{m=1}^{M}\subset\Omega_h$ denote a set of $M$ query points used for a given training step. The training objective is a spatially subsampled mean-squared error over the full trajectory,
\begin{equation}
    \mathcal{L}(\theta) \;=\; \frac{1}{N_T\,M}\sum_{k=0}^{N_T-1}\sum_{m=1}^{M}\, \big\|\, \tilde{q}_\theta(t_k,\xi_m) - q(t_k,\xi_m)\big\|_2^2.
    \label{eq:loss}
\end{equation}
The surrogate inherits physical structure from the latent state evolution in \eqref{eq:latent_euler} and the terrain conditioning of \eqref{eq:cond_decoder}, while the subsampled loss ensures that the gradient remains dominated by active flood zones rather than dry cells.

\paragraph{Spatial subsampling.}
For the Illinois dataset, propagating gradients through the decoder at all $|\Omega_h|=4{,}188{,}840$ active watershed cells for every time step of every trajectory is infeasible on commodity GPUs. We therefore draw $M\ll |\Omega_h|$ query points per training step, resampled at every minibatch. We employ uniform random sampling across all valid in-domain cells, defined as those whose maximum water depth over time (across any training trajectory) exceeds $0.1$~m, resulting in $1{,}408{,}587$ valid union cells.
Unless otherwise stated, we use uniform sampling with $M=7\times 10^4$ on Illinois and $M=2\times 10^4$ on Texas, distributed as $3.5\times 10^4$ and $1\times 10^4$ points per GPU across a dual-GPU setup on a single compute node.
Critically, the latent state $\{s_{t_k}\}$ is shared across all query points within a step, so the cost of the dynamics rollout is amortized over the full sampled batch.

\paragraph{Input/output normalization.}
Only the rainfall forcing is standardized to zero mean and unit variance before entering the network; the outputs $(h, hu, hv)$ are trained and evaluated in their original physical units. The VAE--ConvLSTM and FNO baselines follow the same convention so that all error metrics reported in Section~\ref{sec:results} are directly comparable across methods.

\subsection{Inference and evaluation metrics}
\label{sec:metrics}

At inference time, the latent state is rolled out once per trajectory via~\eqref{eq:latent_euler} and the decoder~\eqref{eq:cond_decoder} is queried at whatever spatial resolution is required by the downstream task. Because the decoder is meshless, the same trained model can be evaluated on a coarse grid for fast visualization, on the native $30\,\mathrm{m}$ simulation grid for validation against SynxFlow, or at off-grid coordinates such as exact USGS gauge locations that need not lie on the raster. We exercise this off-grid capability empirically in Section~\ref{sec:hydrograph_results} by querying the trained model directly at the recorded $(\mathrm{lon},\mathrm{lat})$ of each gauge. Generalization to query resolutions finer than the training grid (super-resolution evaluation) is supported by the architecture. The off-grid query capability is structural (not shared by VAE--ConvLSTM, or FNO baselines).

The latent state $\{s_{t_k}\}$ acts as a compact dynamical state of the basin, while the meshless decoder provides a flexible observation operator that maps this state to arbitrary sensor or forecast locations. The implications of this property for online data assimilation and operational deployment are revisited in the limitations and outlook (Sections~\ref{sec:limitations}--\ref{sec:outlook}).

We assess the model performance along four complementary axes.
\begin{enumerate}
    \item \textit{Aggregate accuracy.} We report the relative root-mean-squared error (rRMSE), averaged over all valid time--space evaluation points, for the full state vector and for each conserved variable separately. For a held-out/test trajectory,
\begin{equation}
    \text{rRMSE}_c \;=\; \frac{\sqrt{\frac{1}{N_T\,|\Omega_{\mathrm{eval}}|}\sum_{k,\xi\in\Omega_{\mathrm{eval}}}\big(\tilde q_c(t_k,\xi)-q_c(t_k,\xi)\big)^2}}{\sqrt{\frac{1}{N_T\,|\Omega_{\mathrm{eval}}|}\sum_{k,\xi\in\Omega_{\mathrm{eval}}}\, q_c(t_k,\xi)^2}},
    \qquad c\in\{h, hu, hv\}.
    \label{eq:rrmse}
\end{equation}
Here $\Omega_{\mathrm{eval}}$ denotes the valid evaluation set: all in-domain cells on Texas and the valid union cells on Illinois.

    \item \textit{Event-level hydrograph metrics at USGS gauges.} At each USGS gauge we additionally compute the Nash--Sutcliffe efficiency (NSE), Kling--Gupta efficiency (KGE), and peak-depth relative error $\varepsilon_{h_{\mathrm{peak}}}$ on the predicted water-depth time series. We evaluate these on water depth rather than water-surface elevation because NSE is invariant to the constant bed-elevation shift while the KGE bias ratio $\beta=\mu_{\mathrm{sim}}/\mu_{\mathrm{obs}}$ is dominated by the $\sim 200\,\mathrm{m}$ bed elevation when computed on WSE. We query the meshless decoder at the exact $(\mathrm{lon},\mathrm{lat})$ of each gauge rather than at the nearest raster cell, exercising a structural capability not shared by grid-based baselines.

    \item \textit{Flood-extent metrics.} For a prescribed depth threshold $\tau$ we convert both predicted and reference depth fields to binary inundation masks $M_\tau(t,\xi)=\mathbf{1}[h(t,\xi)>\tau]$ and report the critical success index (CSI, also known as intersection-over-union), F1, precision, and recall, aggregated over time and space.

    \item \textit{Computational efficiency.} For each model, we report the wall-clock time per predicted trajectory and the peak training memory, and compare against the SynxFlow simulator to quantify the speedup achieved by the learned surrogate.
\end{enumerate}

\section{Results}
\label{sec:results}

This section evaluates the CLDNet of Section~\ref{sec:methodology} on the two datasets introduced in Section~\ref{sec:dataset}. We structure the evaluation along four complementary axes defined in Section~\ref{sec:metrics}: aggregate reconstruction accuracy averaged over valid evaluation points; event-level fidelity at high-water-level sites selected from the SynxFlow reference where peak inundation is most severe; flood-extent agreement under depth thresholds; and computational efficiency relative to the SynxFlow reference. The Illinois reference simulator is separately validated against USGS WSE observations in Section~\ref{sec:illinois_validation}. Throughout, we compare against the unconditional LDNet baseline to isolate the contribution of terrain conditioning, and against two data-driven baselines, a VAE--ConvLSTM model and a FNO model, to position CLDNet against established flood-surrogate paradigms.

\subsection{Experimental setup}
\label{sec:exp_setup}

\paragraph{Architectures and training.}
For both datasets we instantiate CLDNet (Section~\ref{sec:cldnet}) with a dynamics network of depth $8$ and width $50$ and a reconstruction network of depth $10$ and width $300$. The latent dimensionality reflects the complexity of the target dynamics. Specifically, $d_s=30$ suffices on the Texas benchmark with scalar forcing, whereas the spatially heterogeneous Illinois forcing and substantially larger active-cell domain require $d_s=200$. The Fourier-feature frequency count ($m = 10$ for Texas and $m = 32$ for Illinois) of the coordinate embedding $\gamma(\xi)$ is scaled similarly. We tested a range of latent dimensions on both datasets and used the values that gave the best validation rRMSE at fixed wall-clock budget. All models are trained with the Adam optimizer at an initial learning rate of $10^{-3}$, cosine-decayed to $10^{-6}$ over the full schedule, with gradient clipping at norm $1$. Training uses PyTorch Distributed Data Parallel across multiple GPUs on a single node, with spatial decoding carried out in chunks to respect per-GPU memory limits. Spatial subsampling follows Section~\ref{sec:training}: $M=2\times 10^4$ query points per step on Texas and $M=7\times 10^4$ on Illinois, drawn uniformly from in-domain cells unless otherwise noted.

\paragraph{Evaluation.}
Unless otherwise stated, aggregate accuracy is reported as relative root-mean-squared error (rRMSE, Eq.~\ref{eq:rrmse}) in percent, averaged over all valid time--space evaluation points and computed per conserved variable ($h$, $hu$, $hv$) and as an aggregate over all three. On Illinois, rRMSE is evaluated over the wet-union cell set used for spatial sampling, i.e., in-domain cells whose maximum water depth over the training trajectories exceeds $0.1\,\mathrm{m}$ (Section~\ref{sec:training}); flood-extent metrics are computed over the same wet-union mask. Flood-extent metrics use the depth thresholds $\tau\in\{0.1,0.5\}\,\mathrm{m}$. All surrogate errors are relative to SynxFlow, which is itself validated against USGS WSE observations (Section~\ref{sec:illinois_validation}).

\subsection{Baselines}
\label{sec:baselines}

\paragraph{Unconditional LDNet.}
As a direct ablation of the terrain-conditioning mechanism, we train a vanilla LDNet~\eqref{eq:vanilla_decoder} with the same dynamics-network and reconstruction-network architecture, training procedure, and hyperparameters as CLDNet. Any difference in performance between the two models is therefore attributable to the inclusion of the static feature vector in the decoder input.

\paragraph{VAE--ConvLSTM.}
We adapt the encoder--decoder architecture of~\cite{rombach2022high} to map flow snapshots into a $25\times 25\times 8$ latent grid, and model the latent temporal dynamics with a Convolutional LSTM~\cite{shi2015convolutional}. The VAE and ConvLSTM are jointly trained with full teacher forcing for $1200$ epochs using the SOAP optimizer~\cite{vyas2024soap} at learning rate $3\times 10^{-3}$, momentum $(\beta_1,\beta_2)=(0.95,0.95)$, weight decay $0.01$, and a step LR scheduler that decays by $0.1$ every $100$ steps; the preconditioner is updated every $10$ steps. The ConvLSTM is then fine-tuned on latent trajectories alone for $300$ epochs at an initial learning rate of $10^{-4}$, with a teacher-forcing ratio decayed to zero so that the final $100$ epochs are fully autoregressive.

\paragraph{Fourier Neural Operator.}
We train an FNO autoregressively with a 10-step rollout horizon to stabilize training and control error accumulation. We find that using a short rollout horizon (e.g., 1 step) can cause error blow-up in autoregressive prediction on test data, even though the training loss decays nicely. Scalar precipitation is z-score normalized and replicated as a spatially constant channel. The coordinates $(x,y)$ of grid points are linearly scaled to $[0,1]^2$. The inputs are $(x, y, b, \partial_x b, \partial_y b, r, h, hu, hv)$ and the outputs are the next-step $(h, hu, hv)$. Considering the constraints of GPU memory and computational cost, we configure the architecture with $32\times 32$ Fourier modes, $64$ hidden channels, and $4$ Fourier layers, with $9$ input channels and $3$ output channels. The model is trained using the MSE loss with a batch size of $2$ over $100$ trajectories for $50$ epochs. We use the AdamW optimizer with a learning rate of $10^{-3}$ and a weight decay of $10^{-9}$. The training takes approximately $33$ hours. The FNO baseline is trained with a modest hyperparameter sweep over the number of Fourier modes $\{16, 32\}$, hidden channels $\{32, 64\}$, and rollout horizon $\{1, 3, 5, 10\}$.  

\paragraph{Resource comparison.}
Table~\ref{tab:model_comparison} compares model size, total peak training memory, training time, and per-trajectory inference time. Inference time is measured on the same single NVIDIA L40S GPU for every model and is directly comparable across rows; training time and total memory mix GPU generations and reflect each method's practical training setup rather than an apples-to-apples architectural cost. On Texas, CLDNet is roughly $17\times$ smaller than VAE--ConvLSTM and about $10\times$ smaller than FNO in parameter count, and its per-trajectory inference is faster than VAE--ConvLSTM ($4.3\,\mathrm{s}$ vs.\ $7.4\,\mathrm{s/traj}$); FNO achieves the fastest Texas inference ($1.0\,\mathrm{s/traj}$). The same CLDNet design is also trained at Illinois scale (last two rows), where VAE--ConvLSTM and FNO are not tractable at our available batch sizes on the hardware we had access to (see Scaling behavior below). For CLDNet/LDNet, training memory is set by the number of spatial query points $M$ rather than by the simulation grid (Section~\ref{sec:training}), so the Texas-to-Illinois jump from $\approx 30.5$ to $\approx 133.7$\,GiB ($\sim 4.4\times$) primarily reflects our choice of $M=2\times 10^4$ vs $M=7\times 10^4$ together with the wider latent state and Fourier embedding required by the spatially heterogeneous Illinois forcing, rather than the $\sim 33.7\times$ embedded-raster expansion itself. The architecture therefore decouples training memory from grid size: $M$ is a free knob the practitioner can tune to match any GPU budget without changing the network. The negligible differences between LDNet and CLDNet within a dataset ($30.4$ vs $30.5$\,GiB on Texas; $133.3$ vs $133.7$\,GiB on Illinois) reflect only the three extra static-conditioning inputs concatenated to each query coordinate; otherwise the dynamics and reconstruction networks are identical.

\begin{table}[!htb]
\centering
\small
\setlength{\tabcolsep}{6pt}
\renewcommand{\arraystretch}{1.15}
\begin{tabular}{lllcccc}
\toprule
\textbf{Dataset} & \textbf{Model} & \textbf{\# parameters} & \textbf{Training} & \textbf{Memory} & \textbf{Training time} & \textbf{Inference time} \\
\midrule
\multirow{4}{*}{Texas}    & VAE--ConvLSTM     & $14{,}557{,}515$ & 1$\times$\,L40S   & $40.8$\,GiB  & $5$\,h  & $7.4$\,s/traj \\
                          & FNO               & $8{,}965{,}059$  & 1$\times$\,A6000  & $35.1$\,GiB  & $33$\,h & $1.0$\,s/traj \\
                          & LDNet             & $849{,}903$      & 2$\times$\,V100   & $30.4$\,GiB  & $12$\,h & $4.3$\,s/traj \\
                          & CLDNet (ours)     & $849{,}923$      & 2$\times$\,V100   & $30.5$\,GiB  & $12$\,h & $4.3$\,s/traj \\
\midrule
\multirow{2}{*}{Illinois} & LDNet             & $956{,}617$      & 2$\times$\,H100   & $133.3$\,GiB & $17$\,h & $28.8$\,s/traj \\
                          & CLDNet (ours)     & $956{,}713$      & 2$\times$\,H100   & $133.7$\,GiB & $17$\,h & $28.8$\,s/traj \\
\bottomrule
\end{tabular}
\caption{Resource comparison. \textbf{Training hardware} lists the GPU configuration used to train each model (FNO at batch size $2$, all other models at batch size $1$); for the two-GPU runs, the latent dynamics rollout is replicated and spatial query points are sharded across GPUs (Section~\ref{sec:training}). \textbf{Memory (total)} reports the aggregate peak GPU memory across all training devices used by that run. \textbf{Inference time} is measured for every model on the \emph{same} single NVIDIA L40S GPU (the SynxFlow reference hardware of Section~\ref{sec:synxflow}) and is therefore directly comparable across rows; the training-time and memory columns mix GPU generations and reflect each method's practical setup rather than an apples-to-apples architectural cost. Texas rows compare VAE--ConvLSTM, FNO, LDNet, and CLDNet on the $250{,}000$-cell domain; the Illinois rows report LDNet/CLDNet on the full $4{,}188{,}840$-active-cell domain (the $5{,}075\times 1{,}661$ embedded raster).}
\label{tab:model_comparison}
\end{table}

\paragraph{Scaling behavior.} Training memory is the binding constraint for the regular-grid baselines, because the activations, optimizer state, and gradient buffers required by backpropagation through a full-resolution convolutional pipeline grow rapidly with grid size. On the Texas $250{,}000$-cell domain, training the FNO baseline at batch size $2$ on a single A6000 ($48$\,GB) consumes $\approx 35.1$\,GiB, and training the VAE--ConvLSTM at batch size $1$ on a single L40S ($48$\,GB) consumes $\approx 40.8$\,GiB --- both saturating the envelope of these single-GPU configurations on the smaller benchmark. Both architectures store activations at full grid resolution and therefore scale at least linearly with the number of grid cells~\cite{tan2019efficientnet}. The Illinois embedded raster ($\approx 8.43\,\mathrm{M}$ cells) is roughly $33.7\times$ larger than the Texas grid; under this linear scaling, FNO training at batch size $2$ would require $\approx 1{,}183$\,GiB of GPU memory (or $\approx 592$\,GiB at batch size $1$), and VAE--ConvLSTM training at batch size $1$ would require $\approx 1.34$\,TiB. These figures are conservative lower bounds, since deeper or wider channels are typically needed to retain accuracy at higher spatial resolution, and they exclude the additional overhead of model and data parallelism. Consequently, neither baseline is tractable at Illinois scale on the hardware available to us, and even a moderate multi-GPU cluster would be insufficient for either model at the batch sizes we use.
By contrast, CLDNet's neural-field decoder decouples training memory from grid size: spatial query points are sharded across GPUs and chunked within each GPU, so the per-iteration memory cost is set by the number of subsampled query points $M$ and the latent/Fourier widths, not by the simulation grid (Section~\ref{sec:training}). The Texas (2$\times$\,V100) and Illinois (2$\times$\,H100) rows of Table~\ref{tab:model_comparison} ($\approx 30.5$ and $\approx 133.7$\,GiB) therefore differ because we deliberately chose $M=2\times 10^4$ vs $M=7\times 10^4$ together with a wider latent state and Fourier embedding to give the dense Illinois domain richer spatial coverage and absorb its spatially heterogeneous forcing --- not because the architecture's memory scales with the $\sim 33.7\times$ raster. With $M$ held at the Texas value, CLDNet could in principle have been trained on Illinois at roughly Texas-level memory at the cost of fewer spatial samples per step. The same knob lets the practitioner match any fixed GPU budget without changing the network, and the two-GPU spatial-point sharding used here extends straightforwardly to more GPUs for larger $M$ or wider networks.

\paragraph{Domain geometry.} Beyond memory, a second, independent constraint is domain geometry. Both VAE--ConvLSTM and FNO operate on a regular Cartesian grid: the convolutional encoder--decoder of VAE--ConvLSTM presupposes a fixed-resolution rectangular tensor, and the FFT at the heart of FNO requires a uniformly sampled tensorial grid. The Texas benchmark fits this assumption because its computational domain coincides with a $250{,}000$-cell rectangular patch. The Illinois domain, by contrast, is geometrically irregular: only $4{,}188{,}840$ of the $\approx 8.43\,\mathrm{M}$ cells in the embedding raster $5{,}075\times 1{,}661$ are in-domain, and the wet-union mask used for evaluation is non-convex and topologically complex, following the Des~Plaines river network and floodplain. Applying VAE--ConvLSTM or FNO would require padding all out-of-domain cells with sentinel values and either masking the loss or accepting predictions on regions that have no physical meaning; the convolution and Fourier kernels still spend representational capacity and compute on those cells. CLDNet's coordinate-based decoder, in contrast, evaluates the field only at in-domain query points, so irregular geometry is handled natively without any masking or padding.

\begin{figure}[!htb]
\centering
\begin{subfigure}[!htb]{0.49\linewidth}
    \centering
    \includegraphics[width=\linewidth]{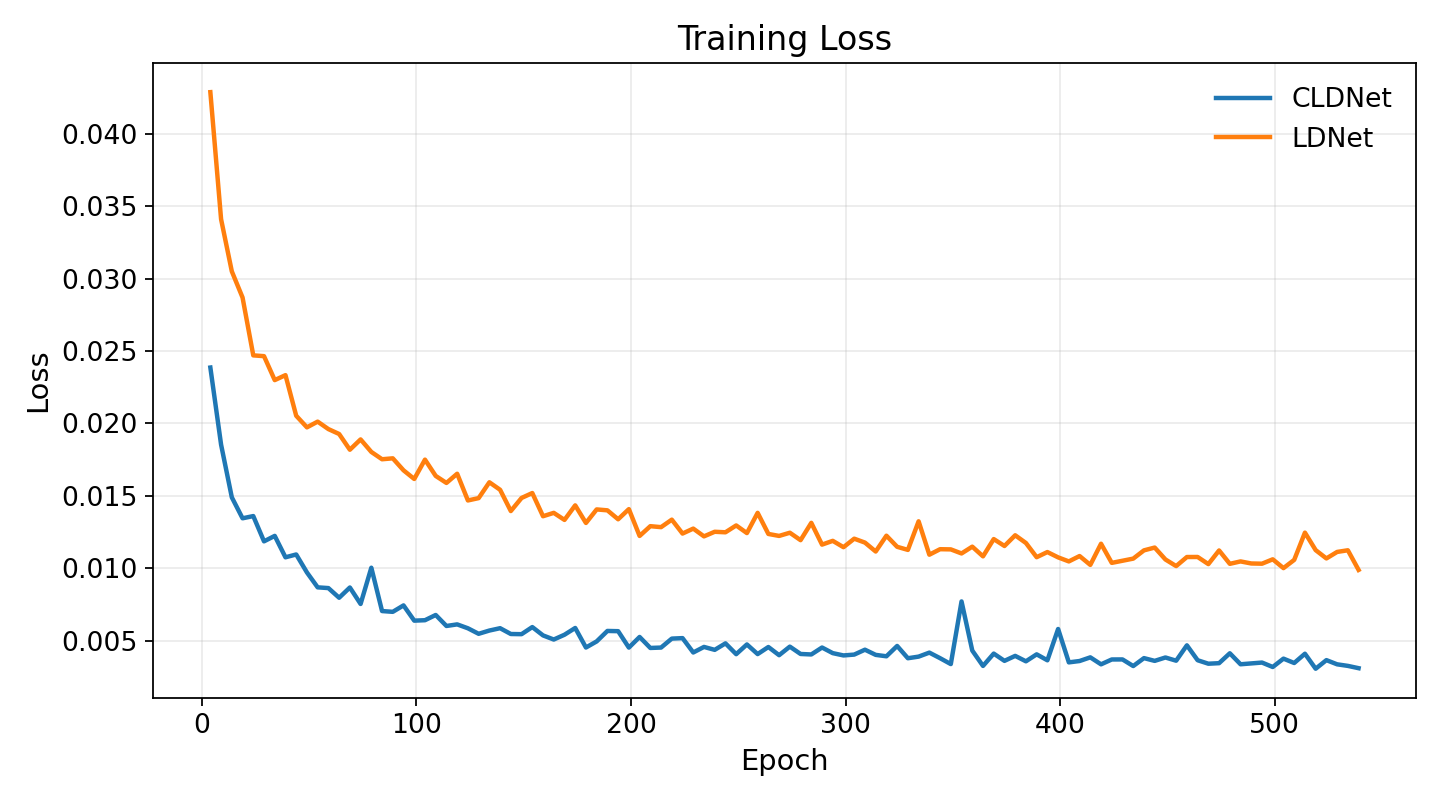}
    \caption{Train loss}
\end{subfigure}
\hfill
\begin{subfigure}[!htb]{0.49\linewidth}
    \centering
    \includegraphics[width=\linewidth]{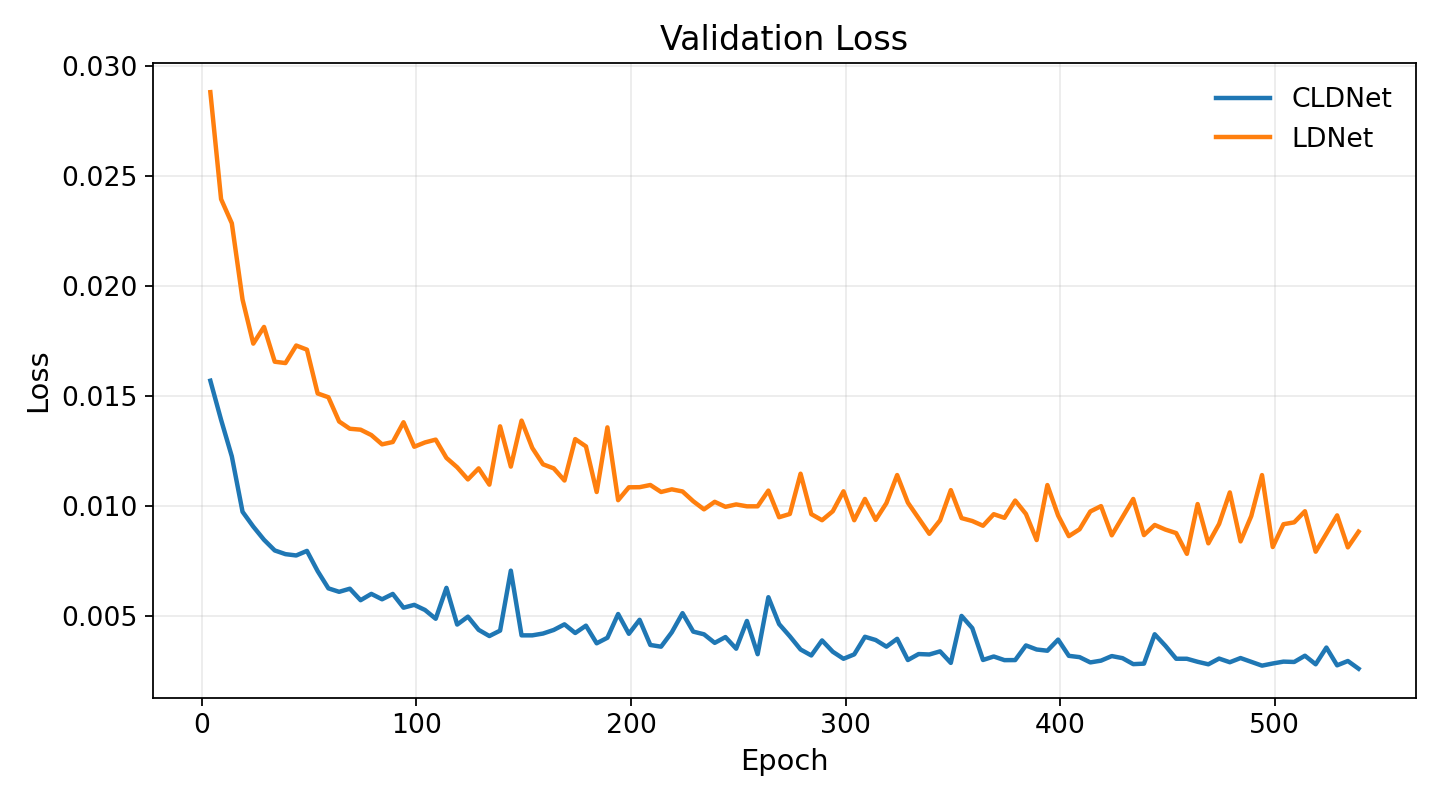}
    \caption{Validation loss}
\end{subfigure}
\caption{Training and validation losses for the  unconditional LDNet and CLDNet on the Des~Plaines dataset. The $x$-axis is the epoch number; the $y$-axis is log-scaled loss on held-out trajectories computed once per $\sim 10$ steps.}
\label{fig:chicago_metrics}
\end{figure}

\paragraph{Training and validation behavior.}
Figure~\ref{fig:chicago_metrics} shows the training and validation losses for the unconditional LDNet and CLDNet on the larger Des~Plaines / Illinois dataset. Both curves decrease steadily and remain well aligned between training and validation, indicating stable optimization without severe overfitting despite the $4{,}188{,}840$-active-cell target. CLDNet achieves significantly lower training and validation losses compared to the LDNet baseline from epoch one, consistent with the design hypothesis that explicit terrain conditioning aids representation. 

\subsection{Aggregate prediction accuracy}
\label{sec:rmse_results}

\subsubsection{Texas benchmark}

Table~\ref{tab:texas_rmse} reports rRMSE on the Texas benchmark. Terrain conditioning yields a $2.15\times$ reduction in the aggregate test rRMSE (from $31.09\%$ for LDNet to $14.44\%$ for CLDNet), with consistent improvements across all three conserved variables. CLDNet also improves aggregate rRMSE by $2.22\times$ relative to VAE--ConvLSTM and by $1.25\times$ relative to FNO. 

\begin{table}[!htb]
\centering
\small
\setlength{\tabcolsep}{6pt}
\renewcommand{\arraystretch}{1.15}
\begin{tabular}{llcccc}
\toprule
 & \textbf{Model} & \textbf{All (\%)} & \textbf{$h$ (\%)} & \textbf{$hu$ (\%)} & \textbf{$hv$ (\%)} \\
\midrule
& VAE--ConvLSTM     & $32.04$          & $26.87$         & $36.33$          & $32.91$ \\
& FNO               & $18.10$          & $12.78$         & $21.51$          & $19.65$ \\
& LDNet             & $31.09$          & $23.72$         & $35.37$          & $33.88$ \\
& CLDNet (ours)     & $\mathbf{14.44}$ & $\mathbf{9.90}$ & $\mathbf{17.17}$ & $\mathbf{15.84}$ \\
\bottomrule
\end{tabular}
\caption{Relative RMSE (\%) on the Texas benchmark, per split and per conserved variable; bold entries are the column best within each split. The discharge errors on $hu$ and $hv$ are systematically larger than those on water depth $h$ because the discharge magnitude vanishes in dry or near-dry cells, so small absolute errors at wet/dry fronts inflate the relative metric.}
\label{tab:texas_rmse}
\end{table}

\subsubsection{Illinois / Des~Plaines case study}

On the Illinois dataset, CLDNet roughly halves the per-trajectory error of the unconditional LDNet on both the average of the held-out test trajectories and the April 17--21, 2013 flood-of-record trajectory used for USGS WSE validation (Table~\ref{tab:traj_rmse}). The improvement is largest for water depth $h$ ($-57\%$ on the test trajectory) and smaller but still substantial for the two discharge components. The absolute rRMSE on the unit-width discharges $(hu, hv)$ remains high ($\sim 40$--$50\%$) on Illinois, reflecting the well-known difficulty of predicting vector fluxes near wet/dry interfaces; we revisit this gap in the limitations (Section~\ref{sec:limitations}).

\begin{table}[!htb]
\centering
\small
\setlength{\tabcolsep}{6pt}
\renewcommand{\arraystretch}{1.15}
\begin{tabular}{llcccc}
\toprule
\textbf{Trajectory} & \textbf{Model} & \textbf{All (\%)} & \textbf{$h$ (\%)} & \textbf{$hu$ (\%)} & \textbf{$hv$ (\%)} \\
\midrule
\multirow{2}{*}{Non-2013 held-out test average ($n=3$)}
& LDNet            & $35.04$          & $32.28$          & $64.44$          & $55.88$ \\
& CLDNet (ours)    & $\mathbf{17.79}$ & $\mathbf{13.47}$ & $\mathbf{48.49}$ & $\mathbf{41.88}$ \\
\midrule
\multirow{2}{*}{2013 USGS-validated event}
& LDNet            & $32.88$          & $30.17$          & $55.96$          & $48.33$ \\
& CLDNet (ours)    & $\mathbf{17.73}$ & $\mathbf{13.31}$ & $\mathbf{43.11}$ & $\mathbf{37.14}$ \\
\bottomrule
\end{tabular}
\caption{Relative RMSE (\%) on the Illinois dataset: the average over the three non-2013 held-out test trajectories, and the April 17--21, 2013 Des~Plaines flood-of-record used for USGS WSE validation in Section~\ref{sec:illinois_validation}. VAE--ConvLSTM and FNO are not included here because their training is not tractable (in memory and computation) at Illinois scale on the hardware available to us.}
\label{tab:traj_rmse}
\end{table}

The conditional architecture delivers the same qualitative benefit across both datasets: roughly halving the error on the metropolitan Des~Plaines case study and cutting it by about two-thirds on the controlled synthetic benchmark. This consistency across regimes is consistent with the design hypothesis of Section~\ref{sec:cldnet} that terrain-derived static features disambiguate spatial heterogeneity that the latent state cannot encode.

\subsection{Event-level evaluation at high-impact flooded sites}
\label{sec:hydrograph_results}

\paragraph{Gauge-level hydrograph metrics.}
At the six Des~Plaines-basin USGS gauges, we compare each surrogate's water-depth time series directly against the USGS observation-derived reference and report standard hydrograph metrics: Nash--Sutcliffe efficiency (NSE), Kling--Gupta efficiency (KGE), and peak-depth relative error $\varepsilon_{h_{\mathrm{peak}}}$. The USGS observations are recorded as stage / WSE; for these depth-based metrics we convert them to gauge water-depth anomalies using the same local datum treatment as in the WSE validation of Section~\ref{sec:illinois_validation}. We work in water depth $h$ rather than WSE because (i)~NSE is invariant to a constant bed-elevation shift applied to both sim and obs, so NSE(depth)~$=$~NSE(WSE) and the two labelings are interchangeable, while (ii)~the KGE bias ratio $\beta=\mu_{\mathrm{sim}}/\mu_{\mathrm{obs}}$ is dominated by the $\sim 200\,\mathrm{m}$ bed elevation when computed on WSE and is only informative when computed on depth, where $\mu$ is of order a few meters. The meshless decoder is queried at the exact $(\mathrm{lon},\mathrm{lat})$ of each USGS gauge rather than at the nearest raster cell, exercising the off-grid-evaluation capability described in Section~\ref{sec:metrics}. Results are expected to be bounded from above by the simulator's own agreement with USGS at the same gauges (Section~\ref{sec:illinois_validation}).

\begin{table}[!htb]
\centering
\small
\setlength{\tabcolsep}{6pt}
\renewcommand{\arraystretch}{1.15}
\begin{tabular}{llccc}
\toprule
\textbf{Gauge} & \textbf{Model} & \textbf{NSE} & \textbf{KGE} & \textbf{$\varepsilon_{h_{\mathrm{peak}}}$ (\%)} \\
\midrule
\multirow{2}{*}{05527800 (mainstem)}
  & LDNet             & 0.608 & 0.794 & $7.26$ \\
  & CLDNet (ours)     & $\mathbf{0.817}$ & $\mathbf{0.858}$ & $\mathbf{4.54}$ \\
\multirow{2}{*}{05528000 (mainstem)}
  & LDNet             & $\mathbf{0.922}$ & 0.865 & $\mathbf{1.39}$ \\
  & CLDNet (ours)     & 0.893 & $\mathbf{0.902}$ & $1.78$ \\
\multirow{2}{*}{05529000 (mainstem)}
  & LDNet             & 0.838 & $\mathbf{0.713}$ & $6.74$ \\
  & CLDNet (ours)     & $\mathbf{0.861}$ & 0.702 & $\mathbf{0.39}$ \\
\multirow{2}{*}{05532500 (mainstem)}
  & LDNet             & $\mathbf{0.543}$ & $\mathbf{0.738}$ & $13.10$ \\
  & CLDNet (ours)     & 0.534 & 0.678 & $\mathbf{2.97}$ \\
\multirow{2}{*}{05531500 (Salt Creek tributary)}
  & LDNet             & $-0.542$ & $\mathbf{0.680}$ & $44.32$ \\
  & CLDNet (ours)     & $\mathbf{0.401}$ & 0.610 & $\mathbf{26.64}$ \\
\multirow{2}{*}{05540130 (W. Br.\ DuPage tributary)}
  & LDNet             & $\mathbf{0.767}$ & 0.783 & $22.29$ \\
  & CLDNet (ours)     & 0.758 & $\mathbf{0.857}$ & $\mathbf{21.69}$ \\
\midrule
\multirow{2}{*}{\textbf{Average}} 
& LDNet         & 0.523 & 0.762 & $15.85$ \\
& CLDNet (ours) & $\mathbf{0.711}$ & $\mathbf{0.768}$ & $\mathbf{9.67}$ \\
\bottomrule
\end{tabular}
\caption{Per-gauge depth-based hydrograph metrics against USGS observation-derived water-depth references for the 2013-04-17 to 2013-04-21 Des~Plaines flood-of-record event, with the bottom rows showing averages over the six gauges. Bold marks the better of LDNet vs.\ CLDNet within each gauge and for the averages.}
\label{tab:hydrograph_metrics_combined}
\end{table}

At individual gauges, the comparison is mixed (Table~\ref{tab:hydrograph_metrics_combined}). CLDNet substantially improves NSE at the two upstream gauges (05527800: $0.608 \rightarrow 0.817$; 05531500 Salt Creek: $-0.542 \rightarrow 0.401$, crossing from unsatisfactory to satisfactory), matches LDNet to within $\sim 0.03$ NSE at three of the four mainstem Des~Plaines gauges, and shows a small regression at 05528000 ($0.922 \rightarrow 0.893$). The mean over the six gauges (NSE $0.523 \rightarrow 0.711$) is dominated by the rescue of 05531500, where the unconditional decoder fails outright. The pattern is consistent with the design hypothesis: terrain conditioning helps most where the unconditional latent state cannot disambiguate morphologically distinct channels, and helps less where LDNet was already accurate. Aggregated over all the gauges, CLDNet improves the mean NSE from $0.523$ to $0.711$ and roughly halves the peak-depth error ($15.85\% \rightarrow 9.67\%$).

\begin{figure}[!htb]
\centering
    \includegraphics[width=\linewidth]{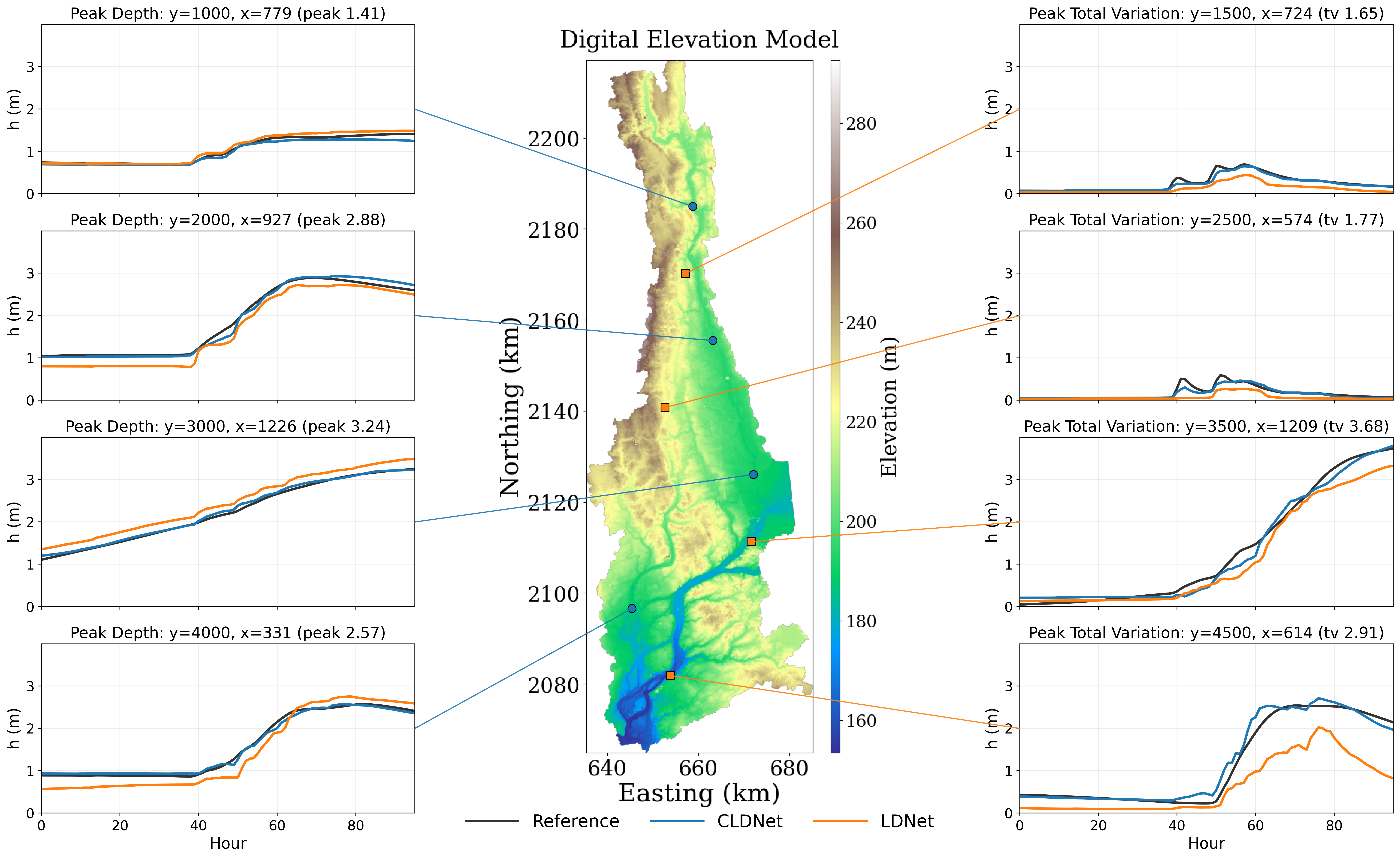}
\caption{Water-depth time series aggregated over high-water-level sites on a representative Des~Plaines test trajectory. The plots compare the baseline LDNet and CLDNet against the SynxFlow reference. Test sites are selected by pre-selecting points at even intervals along the $y$-axis, then identifying the $x$-location with the largest peak depth (left column) and total variation (right column) over time. CLDNet demonstrates significantly improved alignment with the reference.}
\label{fig:depth_hydrograph_sites}
\end{figure}

\paragraph{High-water-level/variation site evaluation.}
Surrogate accuracy matters most where the flood signal is largest. We therefore rank in-domain cells by their peak SynxFlow water depth and the largest depth variation during the 2013 event and select a representative subset of high-water-level sites spread across the basin. Because the decoder is meshless, we can query CLDNet at each selected site without materializing the full grid in memory and compare the predicted depth time series to the SynxFlow reference. Figure~\ref{fig:depth_hydrograph_sites} compares water-depth time series aggregated over the selected sites for a representative Illinois test trajectory: CLDNet tracks both the timing and magnitude of the SynxFlow reference substantially more faithfully than the unconditional LDNet.

\subsection{Flood-extent evaluation}
\label{sec:extent_results}

Continuous-variable rRMSE is complemented by the binary flood-extent metrics defined in Section~\ref{sec:metrics}. Given a depth threshold $\tau$, we convert predicted and reference depth fields into binary inundation masks $M_\tau(t,\xi)=\mathbf{1}[h(t,\xi)>\tau]$ and compare them via the Critical Success Index (CSI, also known as intersection-over-union), F1, precision, and recall. F1 measures overall overlap, precision and recall capture over- and under-prediction biases, and CSI is the operational flood-mapping standard.

Table~\ref{tab:flood_extent_metrics} shows that CLDNet improves the headline operational metrics (CSI and F1) over LDNet across both datasets and both thresholds. The improvement is driven primarily by precision: in every cell of the table, CLDNet's precision is higher than LDNet's, indicating fewer false positives. Recall behavior is mixed: CLDNet's recall increases on Texas at both thresholds and on Illinois at $\tau=0.5\,\mathrm{m}$, but decreases slightly on Illinois at $\tau=0.1\,\mathrm{m}$ ($85.74\% \rightarrow 83.50\%$); this is consistent with CLDNet making slightly more conservative shallow-inundation predictions on the metropolitan domain, with the precision gain ($82.94\% \rightarrow 92.31\%$) more than offsetting the recall cost in CSI/F1. The largest absolute CSI gain is on Texas at $\tau=0.1\,\mathrm{m}$ (CSI from $54.65\%$ to $69.26\%$), where the challenging signal is the detection of thin, terrain-controlled sheet flow. On Illinois at $\tau=0.5\,\mathrm{m}$, CLDNet reaches CSI $=86.04\%$ and F1 $=92.50\%$, corresponding to a well-localized prediction of the severely inundated footprint.

\begin{table}[!htb]
\centering
\small
\setlength{\tabcolsep}{6pt}
\renewcommand{\arraystretch}{1.15}
\begin{tabular}{llccccc}
\toprule
\textbf{Dataset} & \textbf{Model} & \textbf{$\tau$} & \textbf{CSI} & \textbf{F1} & \textbf{Precision} & \textbf{Recall} \\
\midrule
Texas    & LDNet             & $0.1$\,m & $54.65$          & $70.67$          & $80.81$          & $62.79$ \\
Texas    & CLDNet (ours)     & $0.1$\,m & $\mathbf{69.26}$ & $\mathbf{81.84}$ & $\mathbf{96.13}$ & $\mathbf{71.25}$ \\
Texas    & LDNet             & $0.5$\,m & $74.64$          & $85.48$          & $91.77$          & $79.92$ \\
Texas    & CLDNet (ours)     & $0.5$\,m & $\mathbf{82.67}$ & $\mathbf{90.51}$ & $\mathbf{99.20}$ & $\mathbf{83.23}$ \\
\midrule
Illinois & LDNet             & $0.1$\,m & $72.88$          & $84.31$          & $82.94$          & $\mathbf{85.74}$ \\
Illinois & CLDNet (ours)     & $0.1$\,m & $\mathbf{78.07}$ & $\mathbf{87.68}$ & $\mathbf{92.31}$ & $83.50$ \\
Illinois & LDNet             & $0.5$\,m & $74.45$          & $85.35$          & $87.94$          & $82.91$ \\
Illinois & CLDNet (ours)     & $0.5$\,m & $\mathbf{86.04}$ & $\mathbf{92.50}$ & $\mathbf{95.78}$ & $\mathbf{89.43}$ \\
\bottomrule
\end{tabular}
\caption{Flood-extent metrics (\%) at two depth thresholds, for LDNet and CLDNet.}
\label{tab:flood_extent_metrics}
\end{table}

\begin{figure}[!htb]
    \centering

    \begin{subfigure}{0.45\textwidth}
        \centering
        \includegraphics[width=\linewidth]{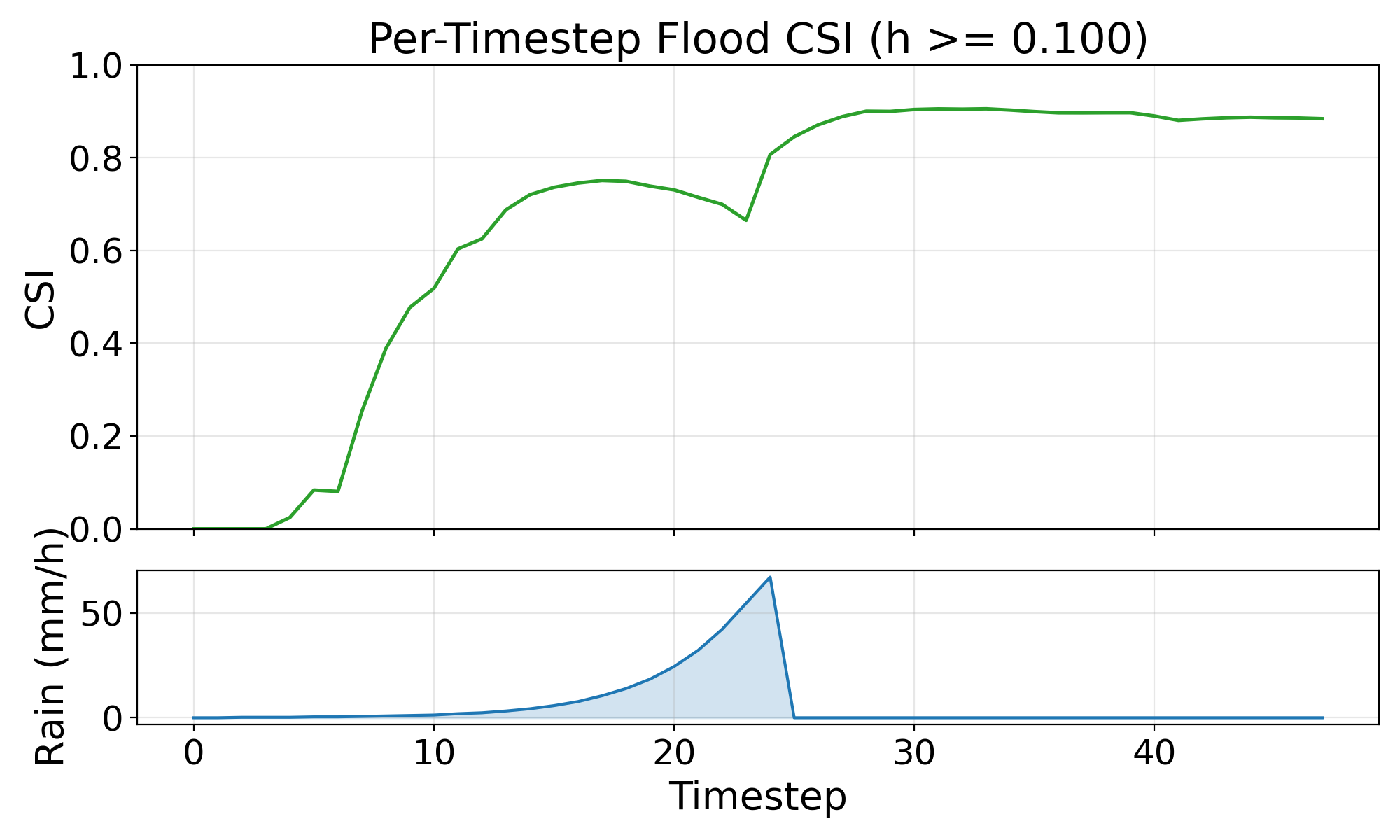}
        \caption{Texas Trajectory}
        \label{fig:texas_csi_curve}
    \end{subfigure}
    \hfill
    \begin{subfigure}{0.45\textwidth}
        \centering
        \includegraphics[width=\linewidth]{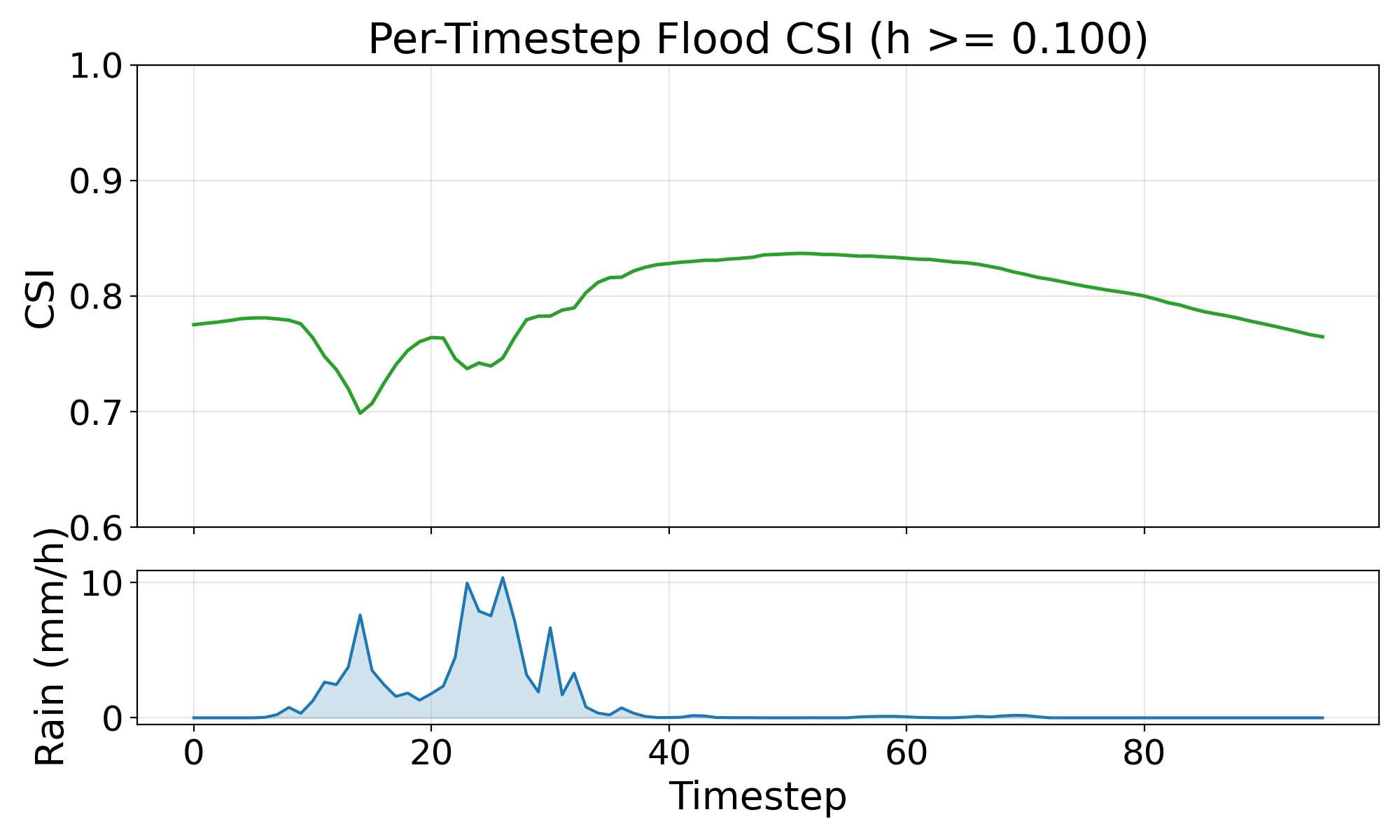}
        \caption{Illinois Trajectory}
        \label{fig:illinois_csi_curve}
    \end{subfigure}
    \caption{Per-timestep CSI for a representative Texas trajectory ($\tau=0.5\,\mathrm{m}$) and a representative Illinois trajectory ($\tau=0.1\,\mathrm{m}$). The lower panel of each subfigure shows the scalar precipitation rate (Texas) or the spatial mean of the precipitation field (Illinois) at each step, so drops in CSI can be read against the rainfall forcing.}
    \label{fig:csi_curves}
\end{figure}

CSI curves in Figure \ref{fig:csi_curves} indicate strong overall model performance, especially when considering the relationship with rainfall-driven dynamics. For the lower threshold ($h\geq 0.1$ m), CSI remains relatively stable in the $\sim 0.75$--$0.85$ range, reflecting good agreement even when including shallow and more uncertain inundation areas. Notably, temporary drops in CSI for the Illinois trajectory align with periods of intense rainfall, suggesting that the model finds it more challenging to capture rapidly evolving flood fronts during storm events; however, performance recovers quickly afterward, indicating robustness during recession and steady-flow phases. Overall, the model performs strongly across time, with only modest degradation during peak rainfall when system dynamics are most complex.
\begin{figure}[!htb]
    \centering
    \begin{subfigure}[b]{0.48\linewidth}
        \centering
        \includegraphics[width=\linewidth]{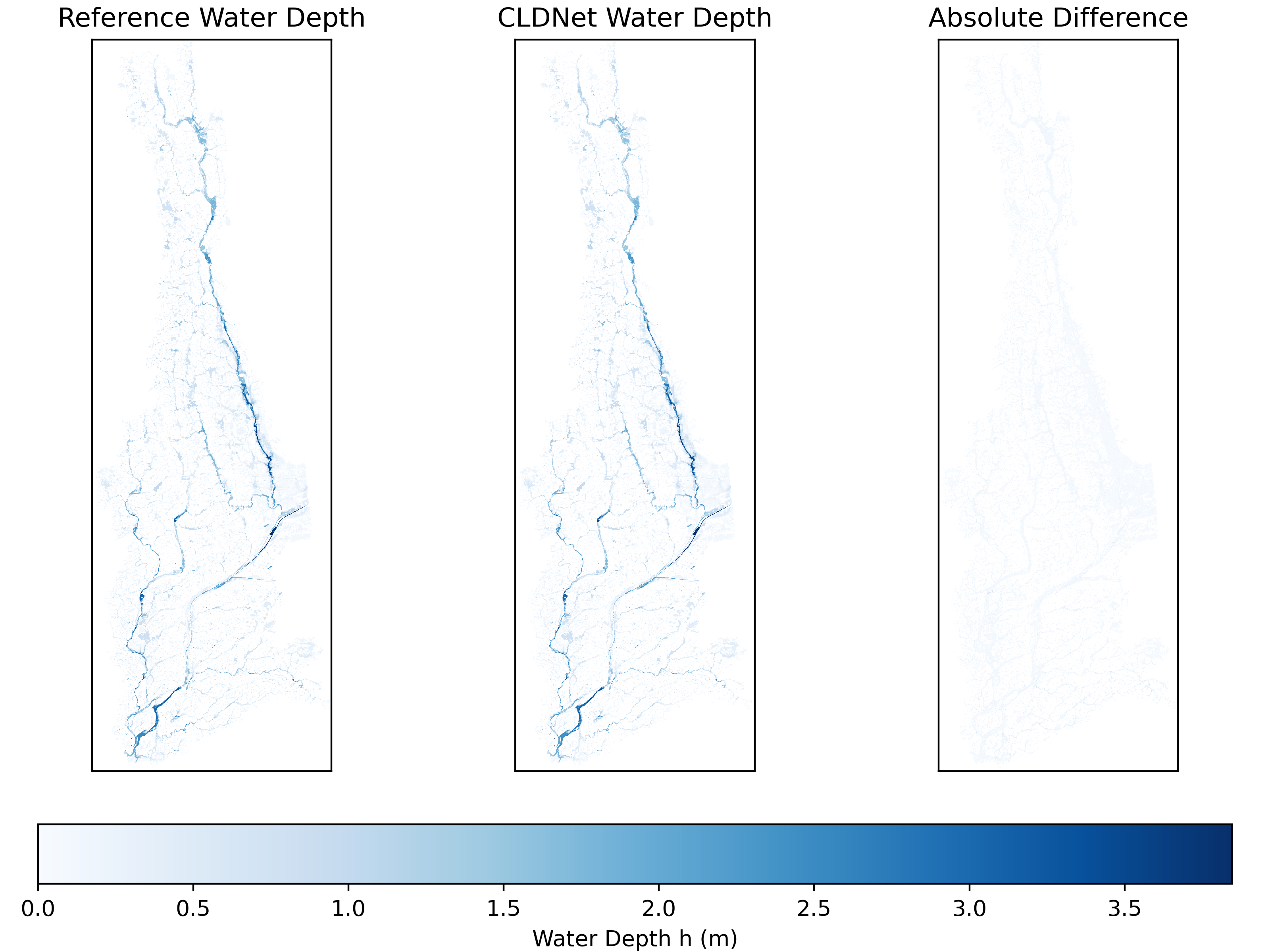}
        \caption{Held-out test trajectory}
        \label{fig:peak_test}
    \end{subfigure}
    \hfill
    \begin{subfigure}[b]{0.48\linewidth}
        \centering
        \includegraphics[width=\linewidth]{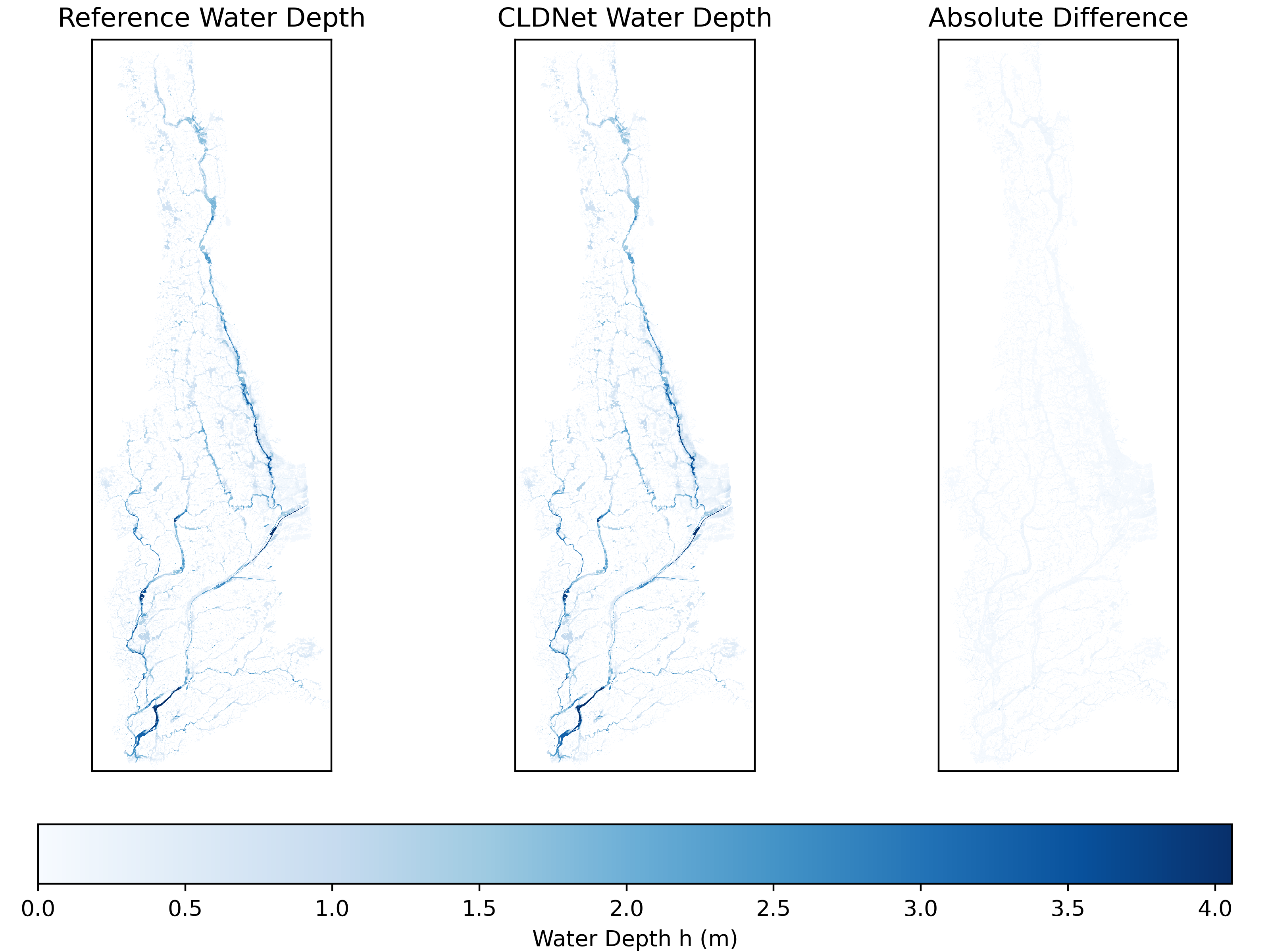}
        \caption{USGS-validated 2013 event}
        \label{fig:peak_2013}
    \end{subfigure}
    \caption{Peak water-depth maps on two representative Des~Plaines / Illinois trajectories. Each subfigure shows three panels: SynxFlow reference (left), CLDNet prediction (center), and absolute error (right).}
    \label{fig:peak_flood_illinois}
\end{figure}

Figure \ref{fig:peak_flood_illinois} shows the peak-inundation maps for the SynxFlow reference and the CLDNet prediction, together with the absolute-error map. The mainstem channel and the major tributary structure align between the two maps, and the overbank floodplain extent is recovered to within the same depth threshold. The absolute-error map is dominated by errors near wet/dry interfaces and at sharp channel banks, where small horizontal misalignments translate into apparent depth differences; errors in channel interiors and on broad floodplains are smaller. Quantitative flood-extent metrics for these maps are reported in Table~\ref{tab:flood_extent_metrics}.

\subsection{Spatial diagnostics at representative flood stages}
\label{sec:spatial_results}

\begin{figure}[!htb]
\centering
\includegraphics[width=0.95\linewidth]{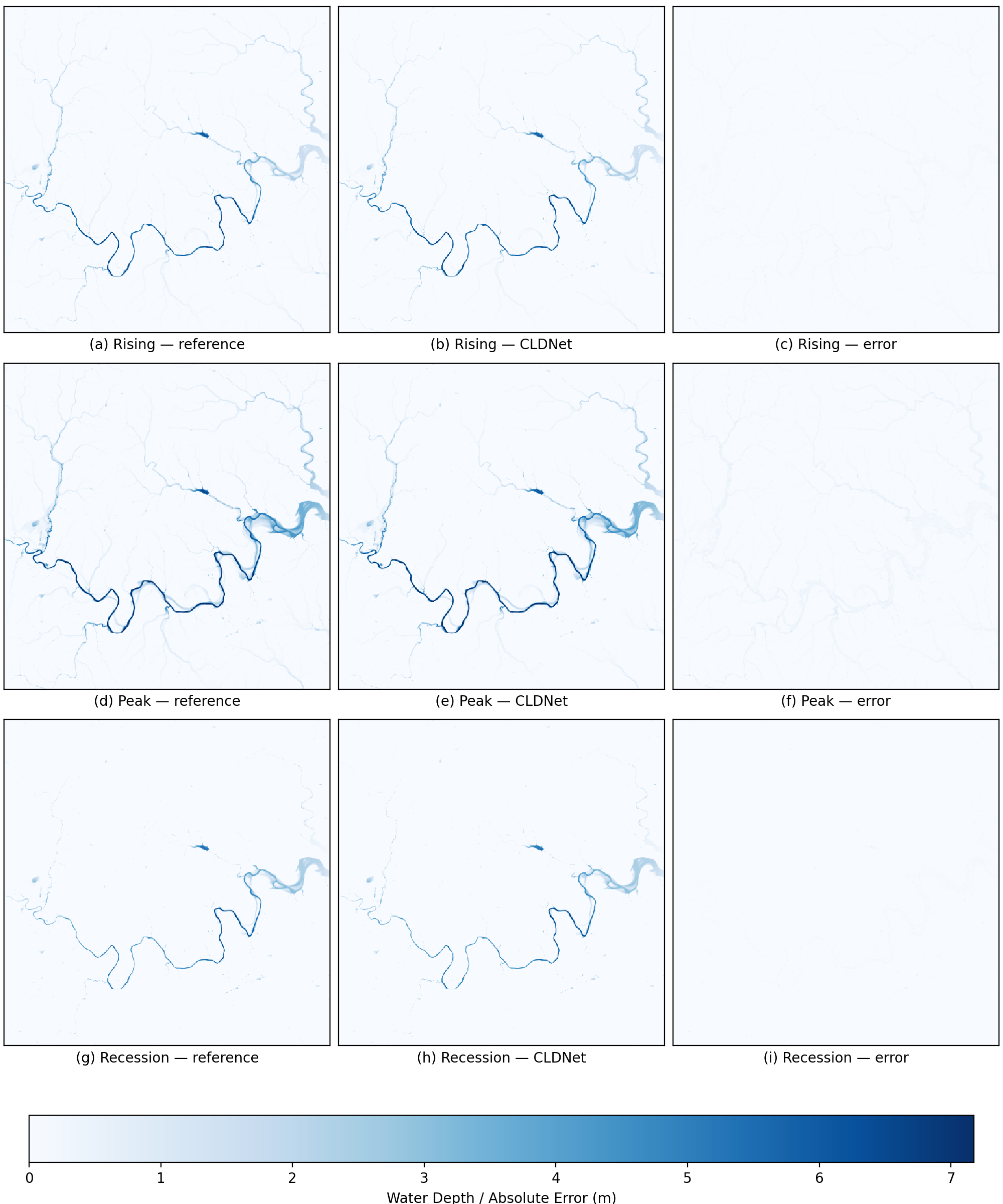}
\caption{Water-depth snapshots at three stages of a representative flood event on the Texas benchmark: rising limb, peak inundation, and recession. Columns show the SynxFlow reference, the CLDNet prediction, and the absolute error.}
\label{fig:spatial_diagnostics_texas}
\end{figure}

\begin{figure}[!htb]
\centering
\begin{subfigure}[b]{0.48\linewidth}
    \centering
    \includegraphics[width=\linewidth]{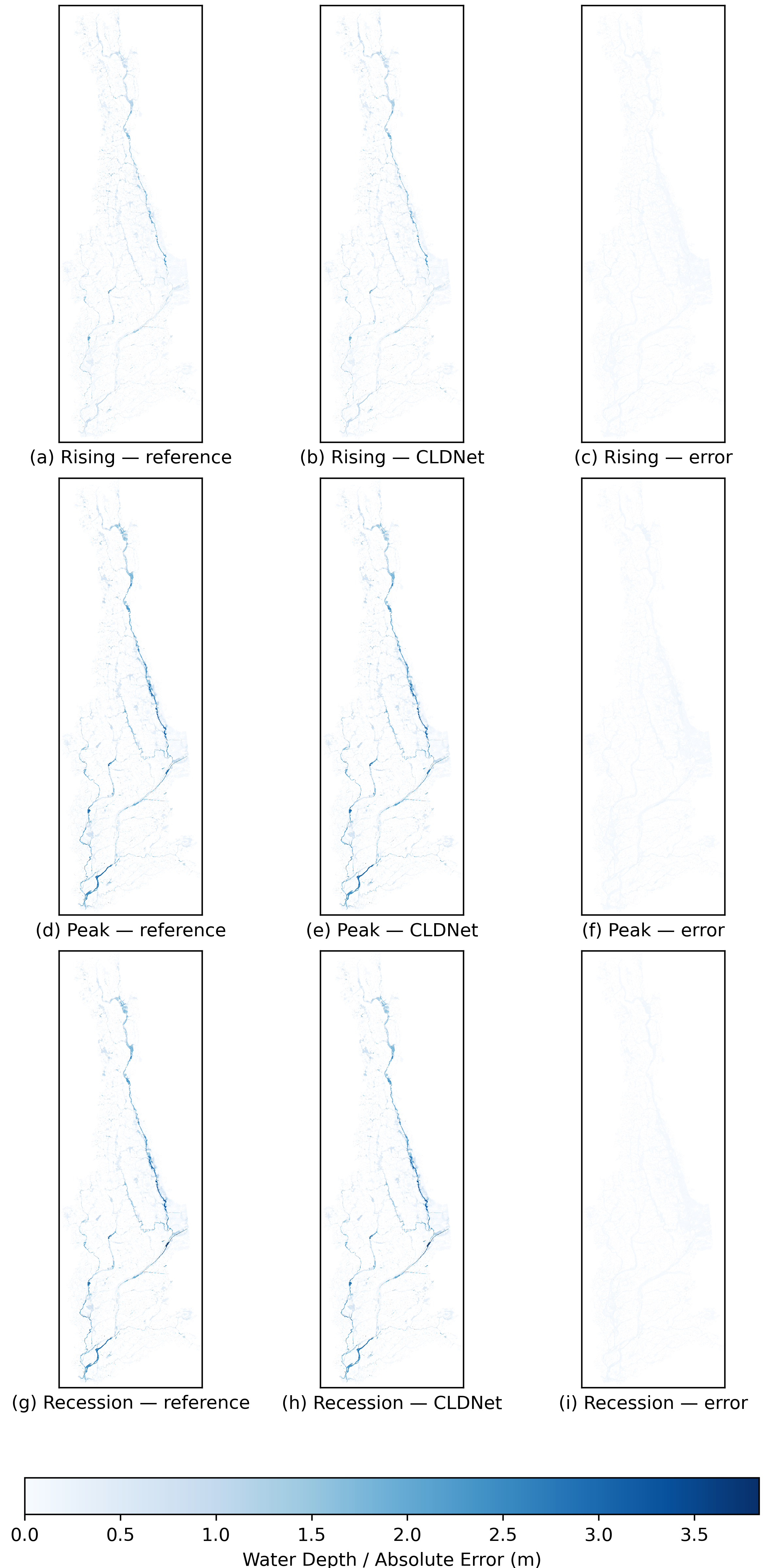}
    \caption{Held-out test trajectory}
    \label{fig:test_flood_snapshot}
\end{subfigure}
\hfill
\begin{subfigure}[b]{0.48\linewidth}
    \centering
    \includegraphics[width=\linewidth]{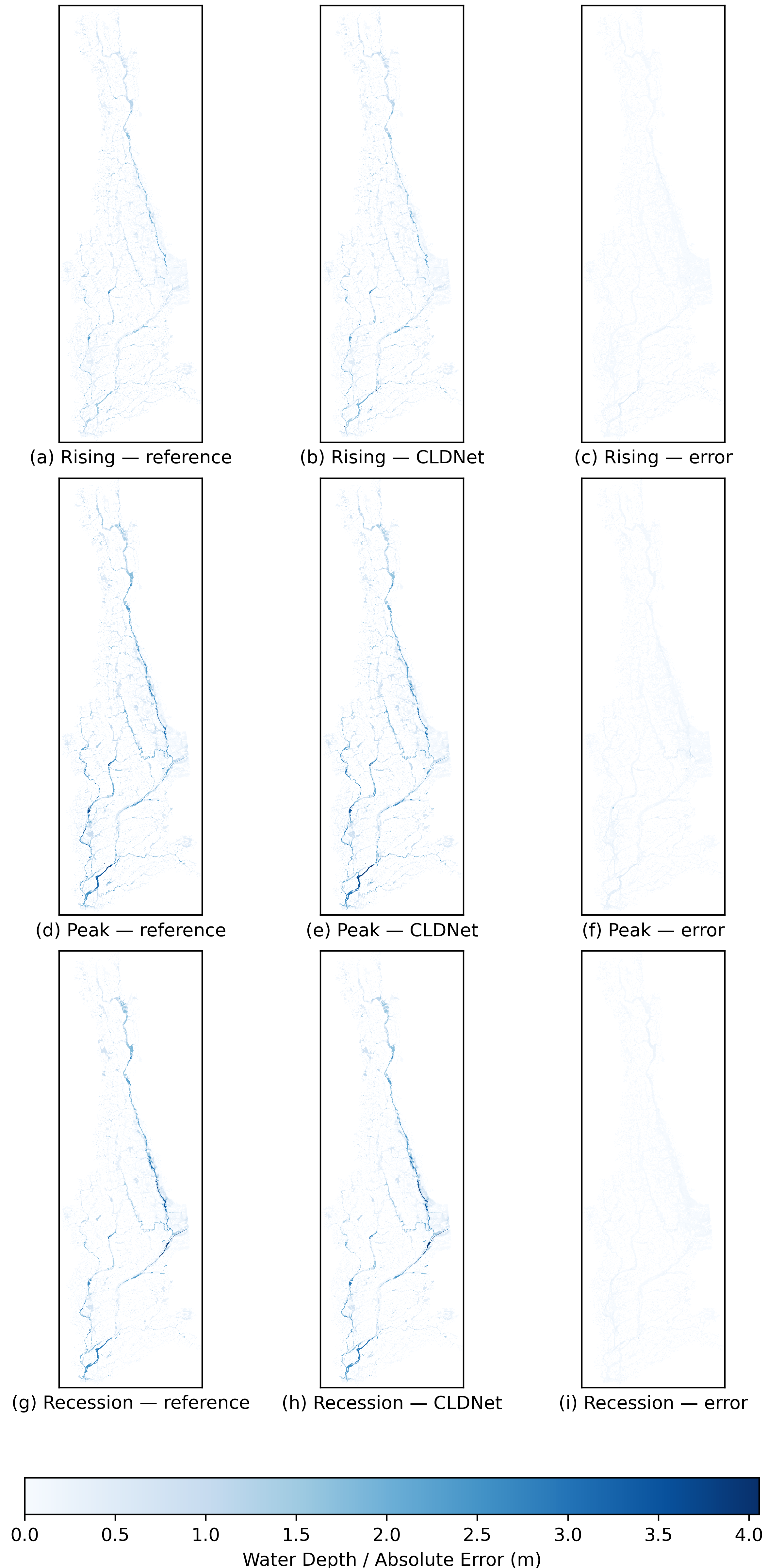}
    \caption{USGS-validated 2013 event}
    \label{fig:usgs_flood_snapshot}
\end{subfigure}

\caption{Water-depth snapshots at three stages (rising limb, peak inundation, recession) of two representative flood events on the Illinois benchmark. In each subfigure the rows correspond to the three stages and the columns show the SynxFlow reference, the CLDNet prediction, and the absolute error.}
\label{fig:spatial_diagnostics_illinois}
\end{figure}

To reveal the spatial distribution of errors rather than only their aggregate magnitude, we plot the reference depth, the CLDNet prediction, and the absolute error at three canonical stages of a flood event, namely the rising limb, the peak inundation, and the recession, for both Texas (Figure~\ref{fig:spatial_diagnostics_texas}) and Illinois (Figure~\ref{fig:spatial_diagnostics_illinois}). These visualizations are designed to reveal (i) whether the model tracks flood-front propagation, (ii) whether errors concentrate near wet/dry interfaces or in channel interiors, and (iii) whether the recession phase is systematically biased.

\begin{figure}[!htb]
    \centering
    \includegraphics[width=0.7\linewidth]{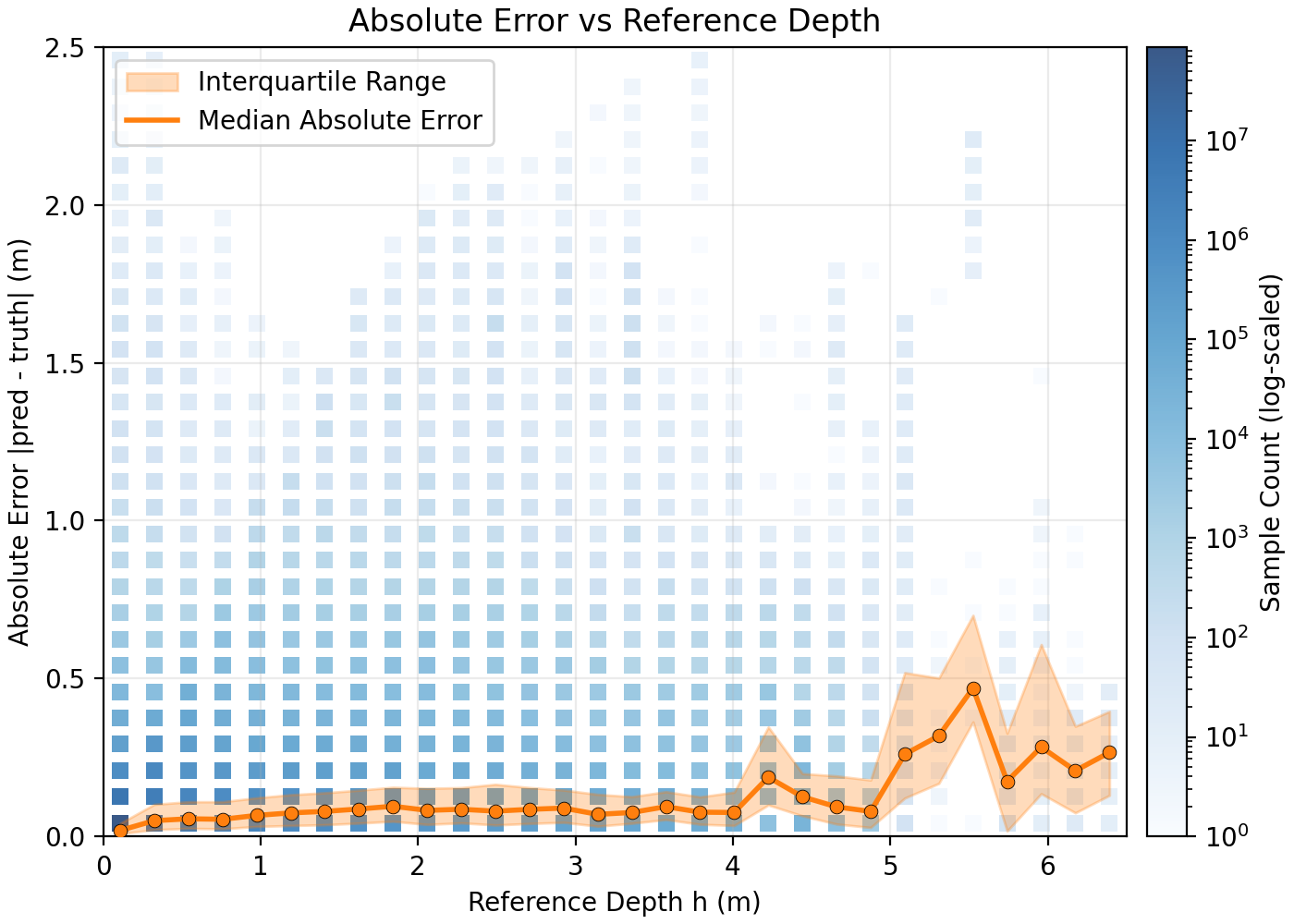}
    \caption{Binned scatter of CLDNet absolute water-depth error $|\tilde h - h|$ versus the SynxFlow reference depth $h$, aggregated over all in-domain cells and all $N_T=97$ snapshots of a representative Des~Plaines / Illinois test trajectory. Color encodes sample count (log-scale, over 7 orders of magnitude). Orange line denotes the median and interquartile range of the absolute error with respect to the reference depth.} 
    \label{fig:binned_scatter_error}
\end{figure}

Figure \ref{fig:binned_scatter_error} shows that the model's absolute error is generally low across all water depths, with most predictions falling below about $0.5\,\mathrm{m}$ error, indicating overall good performance. At shallow depths (roughly $0$--$2\,\mathrm{m}$), where the data is most concentrated, errors are tightly clustered near zero, suggesting the model is both accurate and consistent in these conditions, although occasional large outliers still occur. As depth increases into the mid-range ($2$--$4\,\mathrm{m}$), the spread of errors becomes slightly wider, indicating a modest reduction in reliability. At greater depths ($>4\,\mathrm{m}$), the number of samples drops significantly, and while errors appear more dispersed, the limited data makes firm conclusions difficult; however, it suggests increased uncertainty in deeper water.

\subsection{Computational efficiency}
\label{sec:efficiency}

Computational efficiency is decisive for real-time forecasting and ensemble applications. Table~\ref{tab:trajectory_prediction_time} reports wall-clock time per predicted trajectory for SynxFlow and CLDNet on both datasets. Against the $\sim 55$-minute Illinois SynxFlow trajectory (Section~\ref{sec:illinois_protocol}), CLDNet delivers an end-to-end speedup of $\sim 115\times$, producing a full $96$-hour forecast in tens of seconds at $30\,\mathrm{m}$ resolution over $4{,}188{,}840$ active watershed cells. The corresponding Texas speedup ($\sim 28\times$) is smaller in relative terms because each SynxFlow trajectory there already costs only $\sim 2$ minutes. Training cost is amortized faster on the more expensive simulator: solving the break-even condition $N\cdot(t_{\mathrm{sim}}-t_{\mathrm{inf}}) = t_{\mathrm{train}}$ gives roughly $19$ SynxFlow runs to pay for the Illinois surrogate ($t_{\mathrm{train}}=17$\,h, $t_{\mathrm{sim}}=55$\,min, $t_{\mathrm{inf}}=28.8$\,s), versus about $377$ runs on the cheaper Texas simulator ($t_{\mathrm{train}}=12$\,h, $t_{\mathrm{sim}}=2$\,min, $t_{\mathrm{inf}}=4.3$\,s) --- i.e., the case for surrogate replacement strengthens as the underlying simulator becomes more expensive.

\begin{table}[!htb]
\centering
\small
\setlength{\tabcolsep}{6pt}
\renewcommand{\arraystretch}{1.15}
\begin{tabular}{lcc}
\toprule
\textbf{Method} & \textbf{Texas (s/traj)} & \textbf{Illinois (s/traj)} \\
\midrule
SynxFlow        & $\sim 120$              & $\sim 3300$                \\
CLDNet (ours)   & $4.3$                   & $28.8$                     \\
\midrule
Speedup         & $\sim 28\times$         & $\sim 115\times$           \\
\bottomrule
\end{tabular}
\caption{Wall-clock time per predicted trajectory and end-to-end speedup of CLDNet over SynxFlow.}
\label{tab:trajectory_prediction_time}
\end{table}

\paragraph{What does the speedup unlock?}
The $\sim 115\times$ Illinois speedup is the cost-reduction factor that makes the digital-twin workflows of Section~\ref{sec:introduction} tractable at native metropolitan resolution: repeated rainfall-driven forecast cycles, ensemble forecasting, what-if scenario screening for water management, climate-projection downscaling, and the assimilation cycles that an observation-coupled twin would require. Two concrete budget cases illustrate the implication. A $1000$-member rainfall-forecast ensemble at Des~Plaines requires $\sim 917$ GPU-hours with SynxFlow (roughly \$$730$ at current commercial cloud rates\footnote{Cost estimates assume an AWS on-demand NVIDIA L40S-class instance at $\sim\$0.80$/GPU-hour; they scale proportionally for other providers or on-premises allocations.}), compared with $\sim 8$ GPU-hours with CLDNet (\$$6$); on a single GPU this collapses turnaround from about $38$ days to about $8$ hours --- short enough to fit inside a typical operational forecast cycle. A $30$-year RCP~8.5 climate-downscaling sweep at one representative event per day ($\sim 10{,}000$ runs) decreases from $\sim 9{,}170$ to $\sim 80$ GPU-hours, bringing metropolitan-resolution flood-risk downscaling within the budget of a single research project. More broadly, second-scale inference is the practical latency regime required for dynamic replication in an AI-driven flood digital twin. In a future observation-coupled system, this latency would make it practical to re-run basin-wide forecasts after new precipitation, gauge, or remote-sensing updates, which is the computational prerequisite for the observation-constrained nowcasting outlined in (C5) and Section~\ref{sec:outlook}.

\subsection{Discussion: why does CLDNet outperform the baselines?}
\label{sec:baselines_discussion}

Three observations emerge from the results above. First, the consistent $\sim 2$--$3\times$ rRMSE reduction from LDNet to CLDNet across Texas and Illinois supports the design hypothesis of Section~\ref{sec:cldnet}: at regional scale, terrain heterogeneity is too informative to be recovered implicitly through the latent state and the raw coordinate alone, and explicit conditioning recovers it at negligible parameter cost. Second, the regular-grid baselines (VAE--ConvLSTM and FNO) are inapplicable at Illinois scale on two independent grounds (Section~\ref{sec:exp_setup}, Scaling behavior): they presuppose a Cartesian tensor and cannot natively represent the irregular Des~Plaines domain, and their grid-proportional training memory exceeds single-GPU and even moderate multi-GPU budgets at $\approx 8.43\,\mathrm{M}$ cells. Both points directly illustrate the advantage of the meshless decoder for metropolitan-scale problems. Third, on the Texas benchmark, increasing model parameters (e.g., more Fourier modes or wider channels) may further reduce the error. However, within the explored range, the best configuration ($32$ modes, $64$ channels, horizon $=10$) does not eliminate the gap. Moreover, enlarging the model incurs significantly higher computational cost while yielding only marginal error reduction, suggesting that capacity alone may not resolve the challenge posed by sharp wet/dry interfaces and strongly non-stationary terrain features.

\section{Conclusion}
\label{sec:conclusion}

\subsection{Summary}

We presented the Conditional Latent Dynamics Network (CLDNet) --- a fast, scalable, meshless neural surrogate for the two-dimensional shallow water equations and a candidate building block for AI-driven digital twins of metropolitan flood basins --- that scales regional flood prediction to metropolitan-resolution domains and turns a $\sim 55$-minute, $4{,}188{,}840$-active-cell SynxFlow simulation into a $\sim 29$-second inference. CLDNet factorizes the spatio-temporal flood state into a low-dimensional latent neural ODE driven by rainfall forcing (Section~\ref{sec:ldnet}) and a coordinate-based reconstruction network that decodes the depth and discharge fields at arbitrary query points (Section~\ref{sec:cldnet}). The reconstruction network is explicitly conditioned on a static terrain feature vector comprising standardized elevation, slope magnitude, and Manning roughness, which supplies the spatial heterogeneity that the latent state alone cannot encode at regional scale. Because the decoder is pointwise, training and inference memory are decoupled from the full grid size. This enables a training recipe that includes spatial subsampling and distributed training across events (Section~\ref{sec:training}), making $4{,}188{,}840$-active-cell training tractable on a single dual-GPU node.

We evaluated CLDNet on two complementary datasets: a controlled $250{,}000$-cell Texas benchmark, and a new $4{,}188{,}840$-active-cell Des~Plaines / Illinois case study comprising $114$ real-rainfall storm forcings whose underlying SynxFlow simulator was validated at shape level against USGS WSE observations for the physically observed 2013 flood-of-record (Section~\ref{sec:illinois_validation}). The Illinois domain is roughly $6.9\times$ larger than the largest previously reported active- or area-equivalent flood-surrogate domain. Against the unconditional LDNet baseline, CLDNet reduced aggregate relative RMSE from $31.09\%$ to $14.44\%$ on Texas and from $35.04\%$ to $17.79\%$ on three non-2013 held-out trajectories; on Texas, it also improved aggregate rRMSE by $2.22\times$ relative to VAE--ConvLSTM and $1.25\times$ relative to FNO. CLDNet reached CSI~$=86.04\%$ at the $0.5\,\mathrm{m}$ inundation threshold on Illinois and delivered an end-to-end speedup of $\sim 115\times$ over SynxFlow on the $4.2$M-active-cell domain. The transition to the Illinois / Des~Plaines domain resulted in only a modest increase in training time (from $12$ to $17$ hours, noting that the two runs used different GPU generations as detailed in Table~\ref{tab:model_comparison}), consistent with the meshless coordinate-based architecture's favorable memory scaling as spatial complexity increases by an order of magnitude. Beyond accuracy and speed, the latent state and meshless decoder give CLDNet a compact, gauge-queryable state interface that is structurally compatible with established latent data-assimilation methods, making the architecture a candidate hydrodynamic-replication layer for future AI-driven flood digital twins (Section~\ref{sec:outlook}).

\subsection{Limitations}
\label{sec:limitations}

A few limitations should be kept in mind when interpreting the results and motivate the directions in Section~\ref{sec:outlook}. First, the Illinois experiments are conducted on a single basin (Des~Plaines) under a fixed antecedent flow state shared across all $114$ forecast windows. The surrogate map~\eqref{eq:surrogate} does not currently take an initial condition $q_0$ as input, so transfer to a basin in a different antecedent state requires extending the architecture. The present system also does not assimilate observations online; CLDNet should therefore be viewed as a hydrodynamic surrogate component of a flood digital twin rather than a complete operational twin. Coupling the surrogate to a real-time observation stream requires the data-assimilation extensions discussed in (C5) and Section~\ref{sec:outlook}. Second, the surrogate accuracy metrics are reported against SynxFlow. The SynxFlow validation against USGS at the 2013 flood-of-record is itself shape-level (per-gauge mean-recentered WSE), so any residual datum/bed-elevation bias in the simulator propagates into both training and validation targets. Third, the predicted unit-width discharges $(hu, hv)$ remain less accurate than the predicted water depth $h$, and improving discharge accuracy is the natural target before the surrogate can support a discharge-aware operational flood twin. We report $(hu, hv)$ accuracy explicitly because most published flood surrogates target water depth alone and do not benchmark vector discharges, so this gap cannot yet be situated against a community baseline and remains an open target for future research.

\subsection{Outlook}
\label{sec:outlook}

Several extensions follow naturally from the architecture and results, each of which we view as directly enabled by the scalability and cost reduction demonstrated here but not itself demonstrated in this work.

\begin{itemize}
\item \emph{Probabilistic forecasting} could be obtained by replacing the latent neural ODE with a latent stochastic differential equation or by distilling a score / diffusion prior over the latent state --- an approach already demonstrated for high-dimensional data assimilation in~\cite{si2025latentensf, xiao2026ldensf}. The coordinate-based decoder makes this transition cheap because it admits arbitrary spatial queries with negligible marginal cost.

\item \emph{Observation-constrained nowcasting via latent data assimilation.} Building on contribution (C5), the immediate next experimental step is to couple CLDNet to real-time USGS gauge stages and remote-sensing inundation extents using the latent ensemble score filters~\cite{si2025latentensf, xiao2026ldensf} or differentiable variational filters~\cite{levda2026} that the latent state $s_{t_k}$ is structurally compatible with. Concrete sub-questions include the assimilation cadence (per-snapshot vs.\ per-event), the choice of observation operator (raster cell vs.\ exact gauge $(\mathrm{lon},\mathrm{lat})$), and the trade-off between filter ensemble size and per-step latency for an operational digital twin.

\item \emph{Optimal sensor placement and adaptive observation design.} The same latent state and meshless decoder also enable principled selection of where new gauges, remote-sensing footprints, or temporary deployments would most reduce posterior uncertainty about basin-wide flood state, by casting the question as Bayesian optimal experimental design with a derivative-informed neural-operator surrogate~\cite{go2025dinoboed, go2025lano} or as a sequential-design problem solved by policy-gradient reinforcement learning~\cite{shen2026policy}.

\item \emph{Multi-basin transfer} is natural given the meshless decoder: pretraining on the Illinois dataset and fine-tuning on a new basin with a few tens of simulated events could reduce the simulation budget required to stand up a surrogate for a new jurisdiction.

\item \emph{Operational coupling} with hydrologic upstream models (e.g., NWM or WRF-Hydro streamflow forecasts) and with precipitation nowcasts would close the loop toward an end-to-end forecasting pipeline whose latency is dominated by data ingest rather than simulation.

\end{itemize}

Finally, CLDNet's $\sim 115\times$ simulator speedup lowers the per-run cost of metropolitan flood modeling by two orders of magnitude, which is a prerequisite for the large multi-run workflows, including rainfall-ensemble forecasting, what-if scenario screening, and climate-projection downscaling, that currently limit metropolitan flood-risk analysis; demonstrating such workflows end-to-end is a natural next step.

\subsection*{Data and code availability}
We will release a reproducibility package alongside the published version of this paper, comprising: (i)~the Des~Plaines / Illinois dataset ($114$ real-rainfall storm forcings at $30\,\mathrm{m}$ resolution, including the 2013 flood-of-record used for USGS WSE validation), deposited on Zenodo with a DOI; (ii)~preprocessing scripts for assembling the DEM, NLCD-derived Manning field, Stage~IV precipitation, and USGS gauge stage data; (iii)~the exact $100$/$10$/$4$ training/validation/test trajectory split, including the index of the 2013 event within the test split; (iv)~training and inference code, model configuration files, and random seeds used for the reported runs; and (v)~trained CLDNet, LDNet, VAE--ConvLSTM, and FNO checkpoints. Code will be released on GitHub under a permissive open-source license. 

\subsection*{Acknowledgments}
This work was supported in part by the Office of Cybersecurity, Energy Security, and Emergency Response, Department of Energy under interagency agreement through the U.S. Department of Energy contract DEAC02-06CH11357, the National Science Foundation under Grant No. 2325631 and 2245111, as well as the 2025 IDEaS + Cloud Hub with support from Microsoft at Georgia Institute of Technology. 

The authors would like to thank Prof.\ Barnali Dixon (University of South Florida), Prof.\ Subhrajit Guhathakurta and Prof.\ John E. Taylor (Georgia Tech), and Dr.\ Guannan Zhang (Oak Ridge National Laboratory) for helpful discussions in the development of this work.

\bibliographystyle{elsarticle-num}
\bibliography{references}

@article{xia2017efficient,
  title={An efficient and stable hydrodynamic model with novel source term discretization schemes for overland flow and flood simulations},
  author={Xia, Xilin and Liang, Qiuhua and Ming, Xiaodong and Hou, Jingming},
  journal={Water resources research},
  volume={53},
  number={5},
  pages={3730--3759},
  year={2017},
  publisher={Wiley Online Library},
  doi={10.1002/2016WR020055}
}

@inproceedings{tan2019efficientnet,
  title={{EfficientNet}: Rethinking model scaling for convolutional neural networks},
  author={Tan, Mingxing and Le, Quoc},
  booktitle={International conference on machine learning},
  pages={6105--6114},
  year={2019},
  organization={PMLR}
}

@article{o2022derivative,
  title={Derivative-informed projected neural networks for high-dimensional parametric maps governed by {PDEs}},
  author={O'Leary-Roseberry, Thomas and Villa, Umberto and Chen, Peng and Ghattas, Omar},
  journal={Computer Methods in Applied Mechanics and Engineering},
  volume={388},
  pages={114199},
  year={2022},
  publisher={Elsevier},
  doi={10.1016/j.cma.2021.114199}
}

@inproceedings{sun2023rapid,
  title={Rapid flood inundation forecast using {F}ourier neural operator},
  author={Sun, Alexander Y and Li, Zhi and Lee, Wonhyun and Huang, Qixing and Scanlon, Bridget R and Dawson, Clint},
  booktitle={Proceedings of the IEEE/CVF International Conference on Computer Vision (ICCV) Workshops},
  pages={3733--3739},
  year={2023}
}

@article{shi2015convolutional,
  title={Convolutional {LSTM} network: A machine learning approach for precipitation nowcasting},
  author={Shi, Xingjian and Chen, Zhourong and Wang, Hao and Yeung, Dit-Yan and Wong, Wai-Kin and Woo, Wang-chun},
  journal={Advances in neural information processing systems},
  volume={28},
  year={2015}
}

@inproceedings{rombach2022high,
  title={High-resolution image synthesis with latent diffusion models},
  author={Rombach, Robin and Blattmann, Andreas and Lorenz, Dominik and Esser, Patrick and Ommer, Bj{\"o}rn},
  booktitle={Proceedings of the IEEE/CVF conference on computer vision and pattern recognition},
  pages={10684--10695},
  year={2022}
}

@article{xia2019full,
  title={A full-scale fluvial flood modelling framework based on a high-performance integrated hydrodynamic modelling system ({HiPIMS})},
  author={Xia, Xilin and Liang, Qiuhua and Ming, Xiaodong},
  journal={Advances in Water Resources},
  volume={132},
  pages={103392},
  year={2019},
  publisher={Elsevier},
  doi={10.1016/j.advwatres.2019.103392}
}

@article{xia2018new,
  title={A new efficient implicit scheme for discretising the stiff friction terms in the shallow water equations},
  author={Xia, Xilin and Liang, Qiuhua},
  journal={Advances in water resources},
  volume={117},
  pages={87--97},
  year={2018},
  publisher={Elsevier},
  doi={10.1016/j.advwatres.2018.05.004}
}

@article{regazzoni2024learning,
  title={Learning the intrinsic dynamics of spatio-temporal processes through latent dynamics networks},
  author={Regazzoni, Francesco and Pagani, Stefano and Salvador, Matteo and Dede', Luca and Quarteroni, Alfio},
  journal={Nature Communications},
  volume={15},
  pages={1834},
  year={2024},
  doi={10.1038/s41467-024-45323-x}
}

@inproceedings{xiao2026ldensf,
  title={{LD-EnSF}: Synergizing Latent Dynamics with Ensemble Score Filters for Fast Data Assimilation with Sparse Observations},
  author={Xiao, Pengpeng and Si, Phillip and Chen, Peng},
  booktitle={International Conference on Learning Representations (ICLR)},
  year={2026}
}

@inproceedings{tancik2020fourier,
  title={Fourier features let networks learn high frequency functions in low dimensional domains},
  author={Tancik, Matthew and Srinivasan, Pratul P. and Mildenhall, Ben and Fridovich-Keil, Sara and Raghavan, Nithin and Singhal, Utkarsh and Ramamoorthi, Ravi and Barron, Jonathan T. and Ng, Ren},
  booktitle={Advances in Neural Information Processing Systems},
  volume={33},
  pages={7537--7547},
  year={2020}
}

@article{bentivoglio2023rapid,
  author  = {Bentivoglio, Roberto and Isufi, Elvin and Jonkman, Sebastiaan Nicolas and Taormina, Riccardo},
  title   = {Rapid spatio-temporal flood modelling via hydraulics-based graph neural networks},
  journal = {Hydrology and Earth System Sciences},
  volume  = {27},
  number  = {23},
  pages   = {4227--4246},
  year    = {2023},
  doi     = {10.5194/hess-27-4227-2023}
}

@Article{bentivoglio2025multi,
AUTHOR = {Bentivoglio, R. and Isufi, E. and Jonkman, S. N. and Taormina, R.},
TITLE = {Multi-scale hydraulic graph neural networks for flood modelling},
JOURNAL = {Natural Hazards and Earth System Sciences},
VOLUME = {25},
YEAR = {2025},
NUMBER = {1},
PAGES = {335--351},
DOI = {10.5194/nhess-25-335-2025}
}

@inproceedings{kazadi2024pluvial,
  author    = {Kazadi, Arnold N. and Doss-Gollin, James and Silva, Arlei},
  title     = {Pluvial Flood Emulation with Hydraulics-informed Message Passing},
  booktitle = {Proceedings of the 41st International Conference on Machine Learning (ICML)},
  series    = {Proceedings of Machine Learning Research},
  volume    = {235},
  pages     = {23367--23390},
  year      = {2024},
  publisher = {PMLR}
}

@article{lowe2021uflood,
  author  = {L{\"o}we, Roland and B{\"o}hm, Julian and Jensen, David Getreuer and Leandro, Jorge and Rasmussen, S{\o}ren H{\o}jmark},
  title   = {U-{FLOOD} --- Topographic deep learning for predicting urban pluvial flood water depth},
  journal = {Journal of Hydrology},
  volume  = {603},
  pages   = {126898},
  year    = {2021},
  doi     = {10.1016/j.jhydrol.2021.126898}
}

@article{yang2024rapid,
  author  = {Yang, Fang and Ding, Wu and Zhao, Jianshi and Song, Lixiang and Yang, Dawen and Li, Xudong},
  title   = {Rapid urban flood inundation forecasting using a physics-informed deep learning approach},
  journal = {Journal of Hydrology},
  volume  = {643},
  pages   = {131998},
  year    = {2024},
  doi     = {10.1016/j.jhydrol.2024.131998}
}

@article{li2023physical,
  author  = {Li, Changli and Han, Zheng and Li, Yange and Li, Ming and Wang, Weidong and Dou, Jie and Xu, Linrong and Chen, Guangqi},
  title   = {Physical information-fused deep learning model ensembled with a subregion-specific sampling method for predicting flood dynamics},
  journal = {Journal of Hydrology},
  volume  = {620},
  pages   = {129465},
  year    = {2023},
  doi     = {10.1016/j.jhydrol.2023.129465}
}

@article{liu2025comprehensive,
  author  = {Liu, Bo and Li, Yingbing and Ma, Minyuan and Mao, Bojun},
  title   = {A Comprehensive Review of Machine Learning Approaches for Flood Depth Estimation},
  journal = {International Journal of Disaster Risk Science},
  volume  = {16},
  number  = {3},
  pages   = {433--445},
  year    = {2025},
  doi     = {10.1007/s13753-025-00639-0}
}

@article{tellman2021satellite,
  author  = {Tellman, B. and Sullivan, J. A. and Kuhn, C. and Kettner, A. J. and Doyle, C. S. and Brakenridge, G. R. and Erickson, T. A. and Slayback, D. A.},
  title   = {Satellite imaging reveals increased proportion of population exposed to floods},
  journal = {Nature},
  volume  = {596},
  number  = {7870},
  pages   = {80--86},
  year    = {2021},
  doi     = {10.1038/s41586-021-03695-w}
}

@article{rentschler2022flood,
  author  = {Rentschler, Jun and Salhab, Melda and Jafino, Bramka Arga},
  title   = {Flood exposure and poverty in 188 countries},
  journal = {Nature Communications},
  volume  = {13},
  number  = {1},
  pages   = {3527},
  year    = {2022},
  doi     = {10.1038/s41467-022-30727-4}
}

@article{Chegini_2021,
    author = {Chegini, Taher and Li, Hong-Yi and Leung, L. Ruby},
    doi = {10.21105/joss.03175},
    journal = {Journal of Open Source Software},
    month = {10},
    number = {66},
    pages = {3175},
    title = {{HyRiver: Hydroclimate Data Retriever}},
    volume = {6},
    year = {2021}
}

@article{bates2010simple,
  author  = {Bates, Paul D. and Horritt, Matthew S. and Fewtrell, Timothy J.},
  title   = {A simple inertial formulation of the shallow water equations for efficient two-dimensional flood inundation modelling},
  journal = {Journal of Hydrology},
  volume  = {387},
  number  = {1--2},
  pages   = {33--45},
  year    = {2010},
  doi     = {10.1016/j.jhydrol.2010.03.027}
}

@article{bihlo2022physics,
  title={Physics-informed neural networks for the shallow-water equations on the sphere},
  author={Bihlo, Alex and Popovych, Roman O},
  journal={Journal of Computational Physics},
  volume={456},
  pages={111024},
  year={2022},
  publisher={Elsevier},
  doi={10.1016/j.jcp.2022.111024}
}

@article{brecht2025physics,
  title={Physics-informed neural networks for tsunami inundation modeling},
  author={Brecht, R{\"u}diger and Cardoso-Bihlo, Elsa and Bihlo, Alex},
  journal={Journal of Computational Physics},
  volume={536},
  pages={114066},
  year={2025},
  publisher={Elsevier},
  doi={10.1016/j.jcp.2025.114066}
}

@inproceedings{si2025latentensf,
  author        = {Si, Phillip and Chen, Peng},
  title         = {Latent-{EnSF}: {A} latent ensemble score filter for high-dimensional data assimilation with sparse observation data},
  booktitle     = {The Thirteenth International Conference on Learning Representations (ICLR)},
  year          = {2025},
  eprint        = {2409.00127},
  archivePrefix = {arXiv},
  primaryClass  = {cs.LG}
}

@article{levda2026,
  author        = {Si, Phillip and Chen, Peng},
  title         = {{LEVDA}: {L}atent ensemble variational data assimilation via differentiable dynamics},
  journal       = {arXiv preprint arXiv:2602.19406},
  year          = {2026},
  eprint        = {2602.19406},
  archivePrefix = {arXiv},
  primaryClass  = {cs.LG}
}

@article{stride2026,
  author        = {Tong, Yanjie and Chen, Peng},
  title         = {From sparse sensors to continuous fields: {STRIDE} for spatiotemporal reconstruction},
  journal       = {arXiv preprint arXiv:2602.04201},
  year          = {2026},
  eprint        = {2602.04201},
  archivePrefix = {arXiv},
  primaryClass  = {cs.LG}
}

@article{olearyroseberry2024dino,
  author  = {O'Leary-Roseberry, Thomas and Chen, Peng and Villa, Umberto and Ghattas, Omar},
  title   = {Derivative-informed neural operator: {A}n efficient framework for high-dimensional parametric derivative learning},
  journal = {Journal of Computational Physics},
  volume  = {496},
  pages   = {112555},
  year    = {2024},
  doi     = {10.1016/j.jcp.2023.112555}
}

@inproceedings{qiu2024dedeeponet,
  author        = {Qiu, Yuan and Bridges, Nolan and Chen, Peng},
  title         = {Derivative-enhanced Deep Operator Network},
  booktitle     = {Advances in Neural Information Processing Systems (NeurIPS)},
  year          = {2024},
  pages         = {20945--20981},
  eprint        = {2402.19242},
  archivePrefix = {arXiv},
  primaryClass  = {cs.LG},
  doi           = {10.52202/079017-0660}
}

@article{varopl2025,
  author        = {Qiu, Yuan and Dahmen, Wolfgang and Chen, Peng},
  title         = {Variationally correct operator learning: {R}educed basis neural operator with a posteriori error estimation},
  journal       = {arXiv preprint arXiv:2512.21319},
  year          = {2025},
  eprint        = {2512.21319},
  archivePrefix = {arXiv},
  primaryClass  = {math.NA}
}

@article{go2025lano,
  author        = {Go, Jinwoo and Chen, Peng},
  title         = {Sequential infinite-dimensional {B}ayesian optimal experimental design with derivative-informed latent attention neural operator},
  journal       = {Journal of Computational Physics},
  volume        = {532},
  pages         = {113976},
  year          = {2025},
  eprint        = {2409.09141},
  archivePrefix = {arXiv},
  doi           = {10.1016/j.jcp.2025.113976}
}

@article{go2025dinoboed,
  author        = {Go, Jinwoo and Chen, Peng},
  title         = {Accurate, scalable, and efficient {B}ayesian optimal experimental design with derivative-informed neural operators},
  journal       = {Computer Methods in Applied Mechanics and Engineering},
  volume        = {438},
  pages         = {117845},
  year          = {2025},
  doi           = {10.1016/j.cma.2025.117845}
}

@article{shen2026policy,
  author        = {Shen, Kaichen and Chen, Peng},
  title         = {Sequential {B}ayesian optimal experimental design in infinite dimensions via policy gradient reinforcement learning},
  journal       = {arXiv preprint arXiv:2601.05868},
  year          = {2026},
  eprint        = {2601.05868},
  archivePrefix = {arXiv},
  primaryClass  = {cs.LG}
}

@inproceedings{li2020fourier,
  author        = {Li, Zongyi and Kovachki, Nikola and Azizzadenesheli, Kamyar and Liu, Burigede and Bhattacharya, Kaushik and Stuart, Andrew and Anandkumar, Anima},
  title         = {Fourier Neural Operator for Parametric Partial Differential Equations},
  booktitle     = {International Conference on Learning Representations (ICLR)},
  year          = {2021},
  eprint        = {2010.08895},
  archivePrefix = {arXiv},
  primaryClass  = {cs.LG}
}

@article{lu2021learning,
  author  = {Lu, Lu and Jin, Pengzhan and Pang, Guofei and Zhang, Zhongqiang and Karniadakis, George Em},
  title   = {Learning nonlinear operators via {DeepONet} based on the universal approximation theorem of operators},
  journal = {Nature Machine Intelligence},
  volume  = {3},
  number  = {3},
  pages   = {218--229},
  year    = {2021},
  doi     = {10.1038/s42256-021-00302-5}
}

@article{li2024physics,
  author    = {Li, Zongyi and Zheng, Hongkai and Kovachki, Nikola and Jin, David and Chen, Haoxuan and Liu, Burigede and Azizzadenesheli, Kamyar and Anandkumar, Anima},
  title     = {Physics-Informed Neural Operator for Learning Partial Differential Equations},
  journal   = {ACM/IMS Journal of Data Science},
  volume    = {1},
  number    = {3},
  articleno = {9},
  pages     = {1--27},
  year      = {2024},
  doi       = {10.1145/3648506}
}

@article{vyas2024soap,
  title   = {{SOAP}: Improving and Stabilizing {Shampoo} using {Adam}},
  author  = {Vyas, Nikhil and Morwani, Depen and Zhao, Rosie and Shapira, Itai and Brandfonbrener, David and Janson, Lucas and Kakade, Sham},
  journal = {arXiv preprint arXiv:2409.11321},
  year    = {2024}
}

@article{brocca2024dthydrology,
  author  = {Brocca, Luca and Barbetta, Silvia and Camici, Stefania and Ciabatta, Luca and Dari, Jacopo and Filippucci, Paolo and Massari, Christian and Modanesi, Sara and Tarpanelli, Angelica and others},
  title   = {A Digital Twin of the terrestrial water cycle: a glimpse into the future through high-resolution {E}arth observations},
  journal = {Frontiers in Science},
  volume  = {1},
  pages   = {1190191},
  year    = {2024},
  doi     = {10.3389/fsci.2023.1190191}
}

@article{yang2024digitaltwinriverbasins,
  author  = {Yang, Yifan and Xie, Chen and Fan, Ziwu and Xu, Zhonghou and Melville, Bruce W. and Liu, Guoqing and Hong, Lei},
  title   = {Digital twinning of river basins towards full-scale, sustainable and equitable water management and disaster mitigation},
  journal = {npj Natural Hazards},
  volume  = {1},
  number  = {1},
  pages   = {43},
  year    = {2024},
  doi     = {10.1038/s44304-024-00047-2}
}

@article{rapalo2024floodforecastingdigitaltwin,
  author  = {R{\'a}palo, Luis M. C. and Gomes Jr., Marcus N. and Mendiondo, Eduardo Mario},
  title   = {Developing an open-source flood forecasting system adapted to data-scarce regions: A digital twin coupled with hydrologic-hydrodynamic simulations},
  journal = {Journal of Hydrology},
  volume  = {644},
  pages   = {131929},
  year    = {2024},
  doi     = {10.1016/j.jhydrol.2024.131929}
}

@article{nguyen2025alzette,
  author  = {Nguyen, Thanh Huy and Bhattacharya, Sukriti and Wong, Jefferson S. and Didry, Yoanne and Phan, Duc Long and Tamisier, Thomas and Matgen, Patrick},
  title   = {Towards Digital Twin in Flood Forecasting with Data Assimilation Satellite Earth Observations --- A Proof-of-Concept in the {A}lzette Catchment},
  journal = {arXiv preprint arXiv:2505.08553},
  year    = {2025},
  doi     = {10.48550/arXiv.2505.08553}
}

\end{document}